\documentclass[10pt,twocolumn,letterpaper]{article}

\usepackage[pagenumbers]{cvpr} %
\usepackage{algorithm}
\usepackage{algpseudocode}
\usepackage{multirow}
\usepackage[accsupp]{axessibility}

\newcommand{\dm}[0]{\epsilon_{\theta}}
\newcommand{\xdown}[0]{\mathbf{x}}
\newcommand{\xpad}[0]{\hat{\mathbf{x}}}
\newcommand{\cdscore}[0]{\Delta_\mathcal{C}}
\newcommand{\pixel}[0]{\mathcal{Y}}

\newcommand{\name}{ElasticDiffusion\xspace}

\definecolor{cvprblue}{rgb}{0.21,0.49,0.74}
\usepackage[pagebackref,breaklinks,colorlinks,citecolor=cvprblue]{hyperref}

\def\paperID{*****} %
\def\confName{CVPR}
\def\confYear{2024}

\title{\name: Training-free Arbitrary Size Image Generation \\ through Global-Local Content Separation}

\author{Moayed Haji-Ali\\
\and
Guha Balakrishnan\\
Rice University\\
{\tt\small \{mh155, guha, vicenteor\}@rice.edu}
\and
Vicente Ordonez\\
}

\begin{document}

\twocolumn[{
\renewcommand\twocolumn[1][]{#1}%
\maketitle
\vspace{-0.2in}
\begin{center}
\vspace{-12pt}
    \centering
    \includegraphics[width=\textwidth, height=0.55\textwidth]{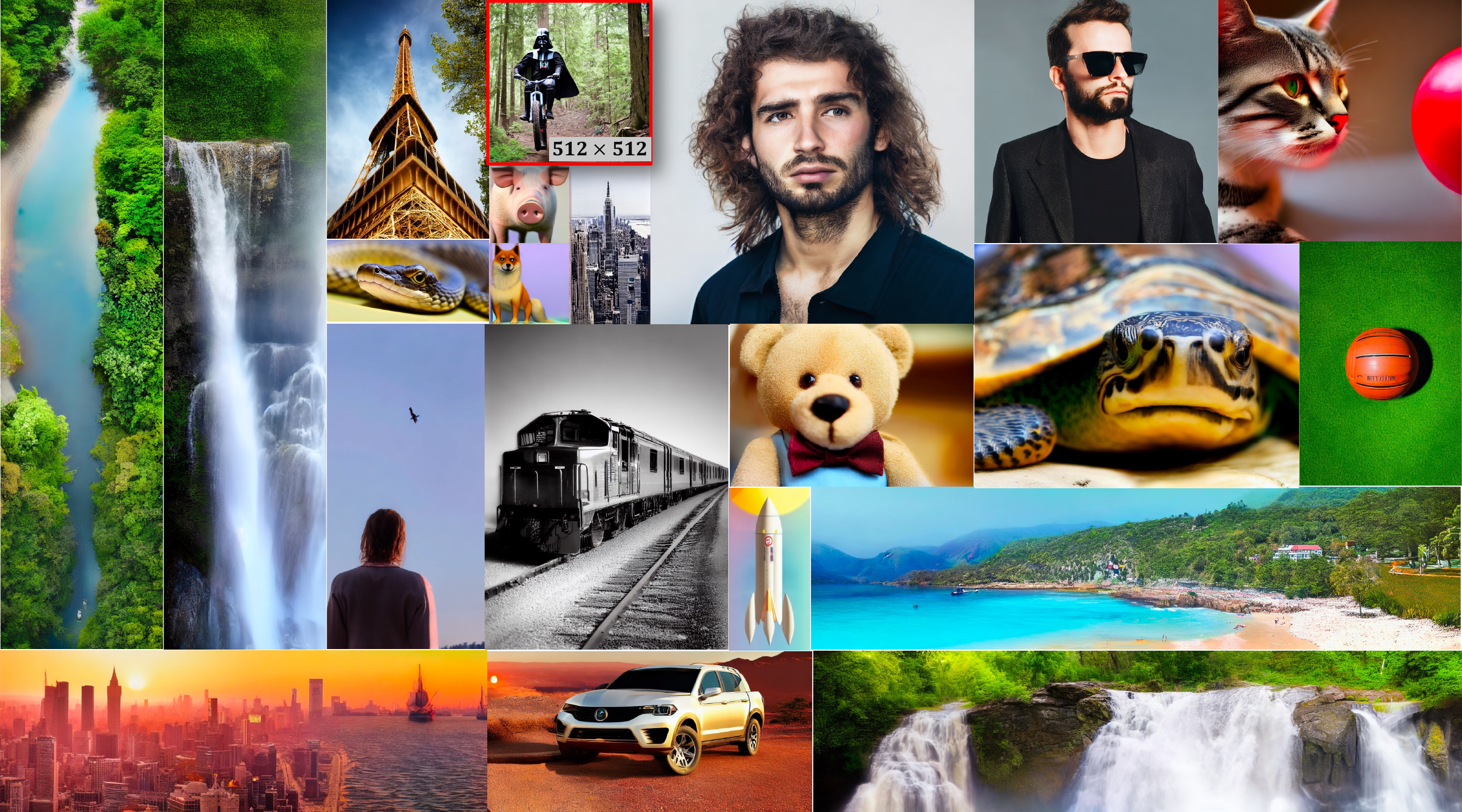}
   \vspace{-0.1in}
\vspace{-0.1in}
    \captionof{figure}{
    \textbf{\name} generates high quality images at arbitrary sizes using a  pretrained diffusion model trained on a single image size, with equivalent memory footprint and no further training. These results are based on $\text{Stable Diffusion}_{1.4}$, which was trained to generate $512 \times 512$ images. %
    The examples shown in this collage are presented without any image cropping, stretching, or post-processing.
    }
    \label{fig:teaser}
\end{center}%
}]

\begin{abstract}
\noindent Diffusion models have revolutionized image generation in recent years, yet they are still limited to a few sizes and aspect ratios. We propose \name, a novel \emph{training-free} decoding method that enables pretrained text-to-image diffusion models to generate images with various sizes. \name attempts to decouple the generation trajectory of a pretrained model into local and global signals. The local signal controls low-level pixel information and can be estimated on local patches, while the global signal is used to maintain overall structural consistency and is estimated with a reference image. We test our method on CelebA-HQ (faces) and LAION-COCO (objects/indoor/outdoor scenes). Our experiments and qualitative 
 results show superior image coherence quality across aspect ratios compared to MultiDiffusion and the standard decoding strategy of Stable Diffusion. Project Webpage: {\href{https://elasticdiffusion.github.io/}{https://elasticdiffusion.github.io/}}
\end{abstract} 
    
\vspace{-0.3in}
\section{Introduction}
\label{sec:intro}
Diffusion models are a powerful family of algorithms that achieve remarkable quality and obtain the current state-of-the-art performance on various image synthesis tasks. As is the case with most deep neural networks, diffusion models are typically trained on one or a few image resolutions. For instance, Stable Diffusion (SD)~\cite{ldm}, one of the most widely adopted diffusion models, is trained on a square images of size 512 x 512, yet fails to maintain its performance at  different aspect ratios during inference time. In practice, many applications require a wide aspect ratio or portrait mode, such as digital billboards, wearable devices, automotive displays, and any application relying on a computer screen. 

Recent studies address the issue of variable image size in different ways. SDXL~\cite{sdxl} and Any-Size-Diffusion~\cite{any-size-diffusion} explicitly finetune models on images with a range of aspect ratios, which requires extensive computation, a quadratic memory footprint, and larger training data. In addition, this strategy requires a set of resolutions to be specified up-front during training time, while the models still struggle to generalize to new resolutions during inference, often resulting in artifacts. Recent works also show remarkable results in generating panoramic images using pretrained diffusion models by overlapping generated patches into a larger image~\cite{multidiffusion, mixture-of-diffusers}. These methods work well for landscape images with repetitive patterns. However, their lack of global guidance limits their abilities to generate images of single objects or faces where global structure is important. Recent work \cite{magicscroll, variable-size-adaptation, anylens} aimed at extending pre-trained diffusion models capabilities to others domains, often in a training-free way \cite{bivdiff, cartoondiff, freecontrol, portraitdiffusion, adaptiveguidance, loco, stylediffusion, conditionvideo, stereodiffusion}. Few concurrent work~\cite{demofusion, cheap-scaling, hidiffusion, large-image, dynamicsr} focus on adapting them to higher resolutions, yet they are constrained to square images.

In this work, we propose \name{}, a novel decoding strategy that takes a pretrained diffusion model and generate images at arbitrary sizes during inference using a constant memory footprint. To achieve this, we revisit the guidance mechanism of conditional diffusion models to decouple global and local content generation. Global content controls the high-level aspects of the image, whereas local content adds finer, more granular details. This separation facilitates generating the local content in patches for images of varying sizes, all while being guided with global content that we derive from a reference image at the diffusion model pretraining resolution. This enables the synthesis of images at diverse resolutions and aspect ratios while adhering the diffusion forward calls to the model's initial training size. To aid in this task, we introduce several techniques including an \emph{efficient} patch fusion method for smooth boundaries, a novel guidance strategy to reduce image artifacts, and a global content resampling technique that amplifies the resolution of diffusion models up to 4X the training size. 

\Cref{fig:teaser} shows a diverse array of images generated with \name{}. Several of theses images include a single object or were generated with extreme aspect ratios, showcasing our method's ability to produce coherent images under various sizes. Quantitatively, \name outperforms baselines across most aspect ratios. More importantly, despite relying on $\text{Stable Diffusion}_{1.4}$, \name obtains comparable FID ($228.87$ vs $230.21$) and CLIP scores ($26.07$ vs $28.06$) as SDXL when generating images at $1024 \times 1024$, which is the native resolution of SDXL.

\section{Related Work}
\label{sec:related}
{\it Diffusion Models} have been widely adopted for their high-quality outputs in generative tasks~\cite{imagen, dalle2, sdxl, ldm, dm-beat-gans,videodm, lvdm,3ddm, control3Diff,make-an-audio, instinpaint}. 
These models involve iterative decoding with many steps leading to high compute and memory requirements.
Recent work addressed these issues by devising faster sampling strategies~\cite{ddim, fastdpm}, hierarchical models~\cite{imagen, dalle2, cascadeddm}, progressive training at increasing resolutions~\cite{glide, ldm, sdxl}, and two-stage models~\cite{sr3, ldm, sdxl, stableSR}. Stable Diffusion (SD)~\cite{ldm} trains a variational auto-encoder to compress images into a low-dimensional $64 \times 64$ latent space and trains a diffusion model on this latent space. To train for higher resolutions, models are initially trained at a $256 \times 256$ resolution, before fine-tuning them at $512 \times 512$ and $1024 \times 1024$ in the case of SDXL~\cite{sdxl}. SD is one of the few large-scale diffusion models that released trained parameters, making it the building block for many subsequent work~\cite{stableSR, prompttoprompt, pix2video, dreambooth, animatelcm, fast-high-res}. However, these models, including SD, are confined to specific resolutions and do not generalize well to aspect ratios unseen during training. Interestingly, despite being presented with a $256 \times 256$ resolution during their early training stages, both SD and SDXL fail to generate realistic images at this resolution after being fine-tuned for larger outputs. \name enables high quality generation at unseen resolutions including re-enabling consistent high quality outputs for SD at $256 \times 256$.

{\it Mixture of diffusers}~\cite{mixture-of-diffusers} and {\it MultiDiffusion}~\cite{multidiffusion} generate panoramic images using a pre-trained diffusion model by generating overlapping crops and combining the generation signal of the overalpping regions. Blending multiple generation signals spatially has been of interest in many previous work \cite{mixture-of-diffusers, blended_diffusion}.
{\it SDXL}~\cite{sdxl} and {\it Any-Size-Diffusion}~\cite{any-size-diffusion} fine-tune a pre-trained SD on a fixed set of resolutions. A concurrent work~\cite{scalecrafter} uses dilated convolution kernels to reduce SD artifacts when generating non-square images. \name{} extends a pre-trained SD to generate images of various sizes at a constant memory requirement and without additional training, all while ensuring global coherence.

{\it Guided diffusion models} devise strategies to condition image generation based on text and other modalities~\cite{cfg, diffusionclip, universal-guidance,audio-token, power-of-sound,classifier-guidance, universal-guidance, ldm, diffusion_posterior_sampling, freedom}. Classifier guidance~\cite{classifier-guidance} uses a pre-trained classifier on noisy images to guide the generation process. Classifier-free guidance~\cite{cfg} eliminates the need for a pretrained classifier but requires training a conditional diffusion model, limiting its applications. Universal Guidance~\cite{universal-guidance} 
applies a pre-trained classifier on the noise-free images produced by DDIM~\cite{ddim}, bypassing training on noisy images. StableSR~\cite{stableSR} uses LoRA~\cite{lora} to condition the generation on low-resolution inputs to achieve superb image super-resolution. Inspired by this, we propose Reduce-Resolution Guidance (\cref{sec:method}) to constrain the generation of high-res images using a lower resolution version, substantially reduceing artifacts without extra training.

\section{Background: Diffusion Models}
\label{sec:background}
A conditional diffusion model \(\epsilon_{\theta} \colon \mathcal{X} \times \mathcal{C} \rightarrow \mathcal{X}\) predicts a less noisy version of the input image \(x \in \mathbb{R}^{H\times W \times 3}\), conditioned on variable \(c \in \mathbb{R}^{D}\) (\eg~a text embedding). Starting with \(x_T \sim \mathcal{N}(0, \mathbf{I})\), the reverse diffusion process progressively denoise $x_T$ over \(T\) steps to generate a realistic image \(x_0\) that conforms to the input condition \(c\) through:
\begin{equation*}
    x_{t-1} = \dm(x_t, c) 
    \text{\qquad for } t = T, T-1, \ldots, 1,
\end{equation*}
where $\epsilon_{\theta}$ is the denoising network. Standard diffusion models operate in the pixel space, but others like Latent Diffusion Models~\cite{ldm} instead operate on a latent image encoding space to reduce memory footprints. A U-Net architecture is commonly used to implement the denoising network and is typically trained on images at a fixed resolution $H \times W$~\cite{dm-beat-gans, ldm, imagen}. The convolutional architecture of a U-Net allows for inputs and outputs of \emph{arbitrary spatial dimensions}. In order to generate an image of a different size $\bar{H} \times \bar{W}$ at inference time, one can sample an initial noise variable $\bar{x}_T \in \mathbb{R}^{\bar{H}\times \bar{W} \times 3}$ and follow the same diffusion process. However, we find that this works poorly in practice, resulting in a significant degradation in output quality.

\textbf{Denoising Diffusion Implicit Models (DDIMs)} introduce a faster non-Markovian sampling strategy, bypassing denoising steps. The reverse diffusion step in DDIM is:

{\begin{equation}
\small
x_{t-1} = {\sqrt{\bar{\alpha}_{t-1}}} \underbrace{\frac{\left( x_t - \sqrt{1 - \bar{\alpha}_t \epsilon_{\theta}^{(t)}(x_t) } \right)}{\sqrt{\bar{\alpha}_t}}}_{\text{predicted }\hat{x}_0^t} + \underbrace{\sqrt{1 - \bar{\alpha}_{t-1}}}_{\parbox{1cm}{ \scriptsize \centering direction pointing to $x_t$}},
\label{eqn:ddim}
\end{equation}}
where $\bar{\alpha}_t = \prod_{i=1}^{t} 1 - \beta_i$ is a cumulative product of the noise levels using a predetermined variance schedule $\beta$. $\hat{x}_0^t$ is a noise-free estimation of ${x}_t$ obtained by subtracting the predicted noise scaled for step $t$. We assume a deterministic sampling process in the DDIM formula~\cite{ddim} for simplicity.

\textbf{Classifier-Free Guidance} updates the reverse diffusion process to condition it on a given condition \( c \) as:

\begin{equation}
\hat{\epsilon}_{\theta}^{(t)}(x_t) = \underbrace{\epsilon_{\theta}^{(t)}(x_t)}_{\parbox{1cm}{\tiny \centering \text{unconditional} \\ \text{score}}} + (1 + w) \cdot \underbrace{(\epsilon_{\theta}^{(t)}(x_t, c) - \epsilon_{\theta}^{(t)}(x_t))}_{\Delta_\mathcal{C}(x,c) \text{: class direction score}}.
\label{eqn:cfg}
\end{equation}
$\epsilon_{\theta}(x, c)$ is a pretrained conditional diffusion model, and $w$ is a scaling factor. The difference between the conditional $\epsilon_{\theta}(x, c)$ and unconditional $\epsilon_{\theta}(x)$ scores, denoted as \emph{class direction score}, gives the guidance direction towards $c$.

\begin{figure}
    \centering   \includegraphics[width=\linewidth]{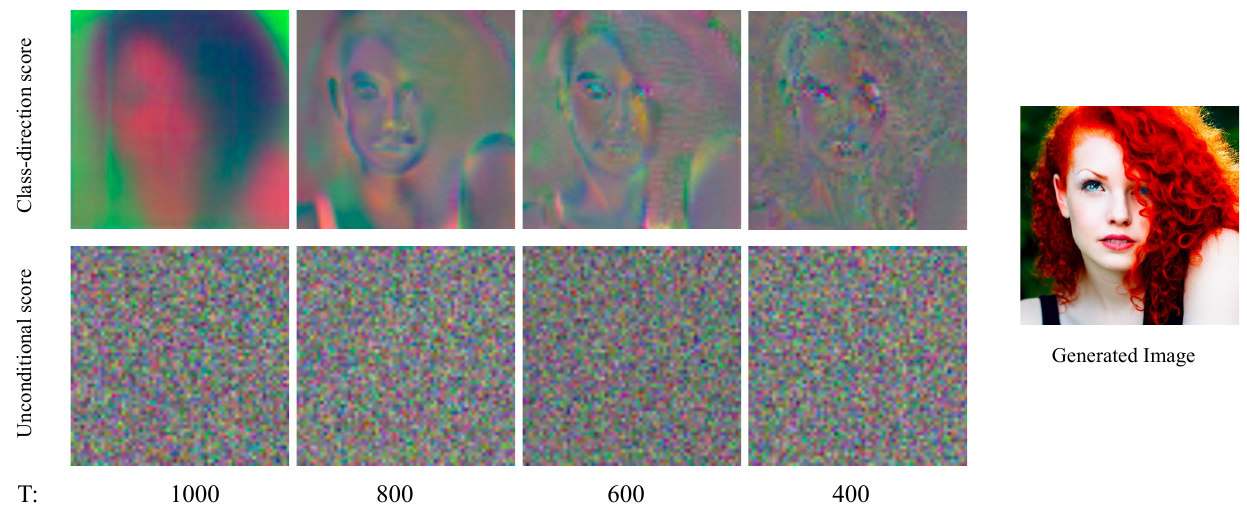}
    \vspace{-2.0em}
    \caption{\textbf{PCA of diffusion scores:} class-direction score (top) dictates global content by clustering on semantic parts, while the unconditional score (bottom) lacks pixel correlations.}
    \label{fig:pca}
    \vspace{-0.2in}
\end{figure}
\vspace{-0.05in}
\begin{figure*}
    \centering
    \includegraphics[width=0.9\linewidth]{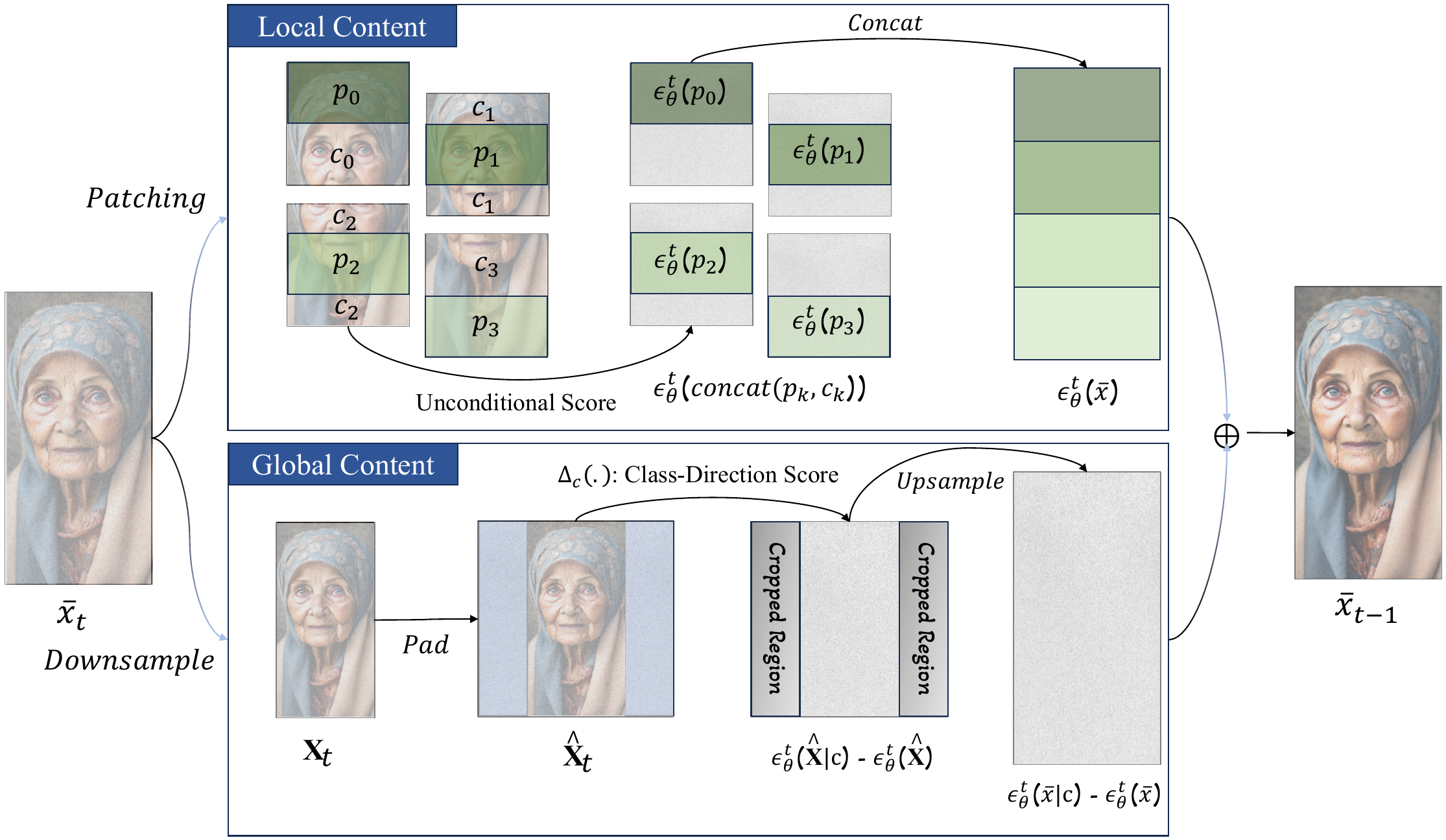}
    \caption{\textbf{Illustration of \name:} We generate images at various sizes by generating local and global content separately. For local content, we partition the latent $\bar{x}_t$ into \emph{non-overlapping} patches $p_k$, each concatenated with context $c_k$ to estimate their unconditional score. For global content, we downsample $\bar{x}_t$ to $\xdown_t$, pad to a square size ($\hat{\xdown}_t$), compute class-direction score ($\Delta_c$), and upscale to match $\bar{x}_t$.}
    \label{fig:rebuttal-illustration}
\end{figure*}
\section{\name}
\label{sec:method}
\noindent The aim of this work is to develop a method capable of synthesizing images at arbitrary size $\bar{H} \times \bar{W}$, %
and conforming to a global condition \(c\) using a pre-trained diffusion model that is limited at inference to its training resolution $H\times W$. To achieve this, we observe a main insight with respect to the diffusion scores in~\cref{eqn:cfg} that we visualize in~\cref{fig:pca}. The class direction score ($\Delta_c$) primarily dictates the image's overall composition by showing clustering along semantic regions (\eg~hair) and edges. Thus, upscaling this score maintains global content in higher resolutions. In contrast, the unconditional score lacks inter-pixel correlation, suggesting a localized influence in enhancing detail at the pixel level. Thus, computing it requires a pixel-level precision. We study the global and local behaviour of these scores in Supp.~\cref{global-local-analysis}. Our intuition behind this observation is that latent pixels, specifically in early diffusion steps where noise is high, display weak and short-range correlations. This leads the unconditional score to focus on local patterns due to these limited pixel interactions. Meanwhile, as $\Delta_c$ represents a latent direction towards a specified global condition, it exerts a global influence by guiding clusters of pixels with similar directions towards semantic regions, overriding their initial noisy state. Leveraging this insight, we propose a method to decouple the generation of local and global content during the diffusion process. Specifically, we compute the unconditional score on local patches of size $H\times W$ while simultaneously resizing a class direction score, originally derived for a reference latent of the dimensions $H\times W$ as well. 
This dual strategy facilitates the generation of images at varied sizes using a pretrained diffusion model at its native resolution, all while maintaining the same memory requirement and without further training. We first present our approach for computing the unconditional score (\cref{subsec:unconditional_score}). We then detail our method to estimate (\cref{subsec:estimation-class-direction}) and upscale the resolution (\cref{subsec:resampling}) of the class direction score. Finally, we combine the two estimated scores with a novel guidance strategy to generate images at arbitrary sizes (\cref{subsec:rrg}). The generation process of \name is illustrated in~\cref{fig:rebuttal-illustration}.
\subsection{Estimating the Unconditional Score}
\label{subsec:unconditional_score}
\begin{figure}[t!]
    \centering
\includegraphics[width=0.95\columnwidth]{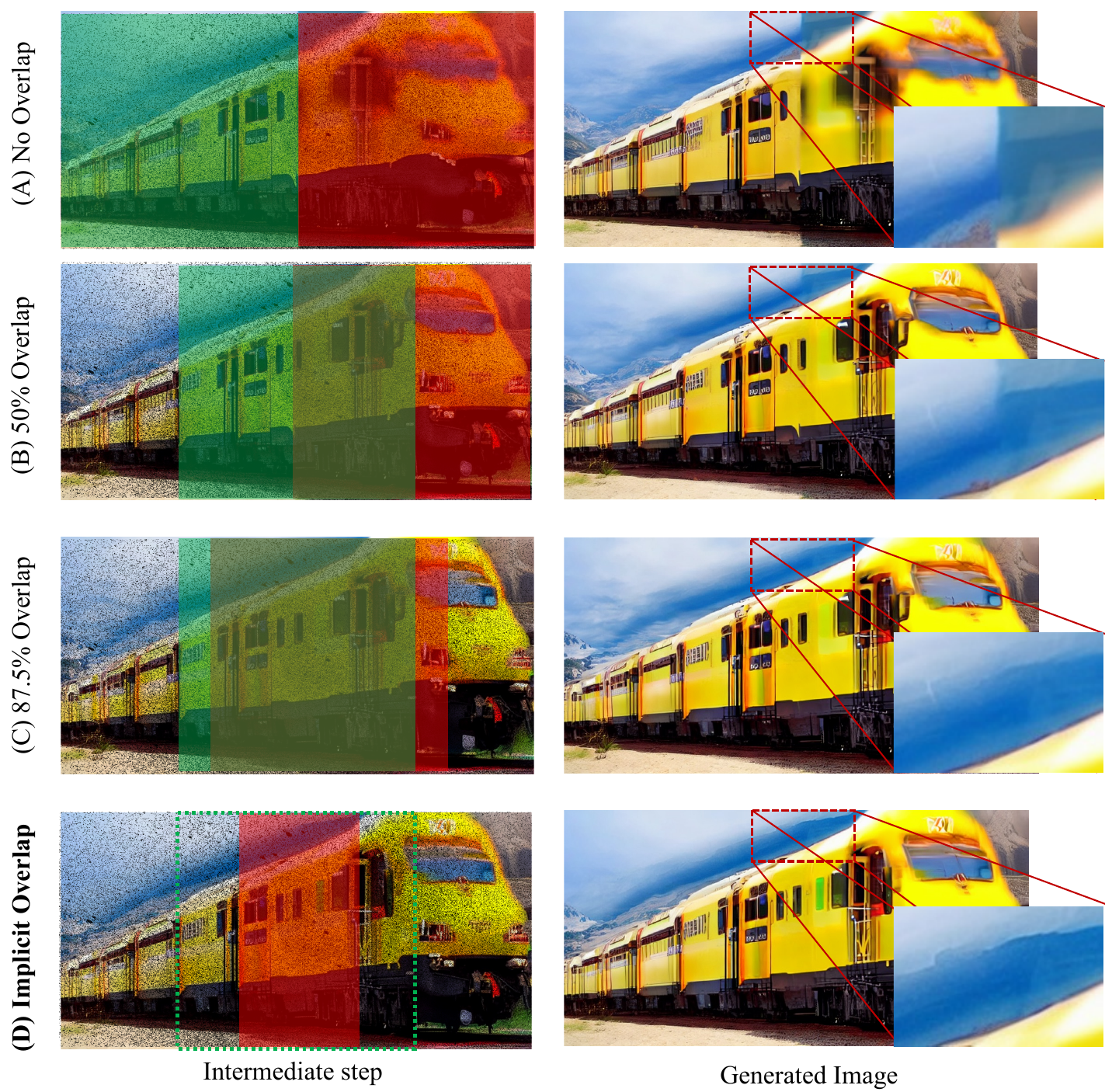}
    \caption{Comparing strategies for calculating diffusion model score on a local patch. No overlap between adjacent patches (A) leads to discontinuities at the boundaries. Strategies (B) and (C), explicitly overlap nearby patches, necessitating substantial overlap to be effective. Our implicit overlapping method (D) achieves superior results with computational demand similar to (B).}
    \label{fig:context-pixs}
    \vspace{-0.2in}
\end{figure}

Building upon previous work \cite{mixture-of-diffusers, stableSR, multidiffusion}, we estimate the unconditional score for a large latent signal \(\bar{x}_t \in \mathbb{R}^{\bar{H}\times \bar{W} \times 3}\) by concatenating scores derived from local patches. Specifically, we divide image $\bar{x}_t$ to $K$ patches $P_k \in \mathbb{R}^{H\times W \times 3}$ and compute the score as $\epsilon_{\theta}(\bar{x}_t) = \{\epsilon_{\theta}(P_k);\; \forall_{k \leq K}\}$. A common challenge encountered with this implementation is discontinuities at boundaries, as illustrated in~\cref{fig:context-pixs} (A). To address this, earlier research calculated the diffusion model score on explicitly overlapping patches and averaged their scores in the intersecting regions~\cite{stableSR, multidiffusion, mixture-of-diffusers}. While this strategy mitigates discontinuities, it requires large overlap between patches, thereby substantially increasing inference time and blurring details. To speed up the process, we introduce a more effective method that enhances boundary transitions in local patches by incorporating contextual information from adjacent patches, thus negating the need for signal averaging in overlapping areas. Specifically, we select patches smaller than the full size $p_k \in \mathbb{R}^{h \times w \times 3}$ with $h < H$ and $w < W$, and concatenate them with context pixels from adjacent patches, denoted as $c_k \in \mathbb{R}^{(H-h)\times (W-w) \times 3}$, to compute the diffusion model unconditional score $\mathbf{S}_u$ as: 
{
\small
\begin{equation}
\label{eqn:uncond-score}
\begin{split}
     \tilde{\epsilon}_{\theta}({x_t})= \{\,\epsilon_{\theta}(p_k \mid c_k)\;| \;\bar{x}_t = \{p_k;\; \forall_{k \leq K}\}
     \},
\end{split}
\end{equation}
}  
where $p_i$ is a local patch and $c_i$ are the context pixels surrounding it. This substantially increases the efficiency of the process. For instance, to generate an image of size $1024 \times 1024$, previous methods~\cite{stableSR, multidiffusion, mixture-of-diffusers} used $87.5\%$ overlap between adjacent patches, necessitating $81$ forward diffusion calls per decoding step. In comparison, \name{} achieves comparable results with only $9$ forward calls, as demonstrated in~\cref{fig:context-pixs} (D). Employing the same number of calls, previous techniques result in obvious boundary discontinuities, as depicted in~\cref{fig:context-pixs} (B). 

\subsection{Estimating the Class Direction Score}
\label{subsec:estimation-class-direction}
A simple approach to estimate a class direction score of an intermediate latent signal $\bar{x}_t \in \mathbb{R}^{\bar{H} \times \bar{W} \times 3}$ is to upsample the score from a reference latent $x_t \in \mathbb{R}^{N \times M \times 3}$ to $\bar{H} \times \bar{W}$. This is possible due to our observation that the class direction score represents a latent direction that can be shared between nearby pixels. We validate this observation empirically in Supp.~\cref{global-local-analysis}. We choose $N < H \textit{ and } M < W$ such that $N \times M$ is as close as possible to $H \times W$ and $\frac{\bar{H}}{\bar{W}} = \frac{N}{M}$. This is important to maintain the aspect ratio and prevent stretching the global content. Formally, we compute the class direction score $\mathbf{S}_d$ as:
\begin{equation}
\begin{split}
    \xdown_t &\leftarrow \mathrm{Downsample}(\bar{x}_t, N \times M), \\
    \cdscore(\bar{x}_t, c) &= \mathrm{Upsample}(\Delta_\mathcal{C}(\mathbf{x}_t, c), \bar{H} \times \bar{W}),
\end{split}
\label{eqn:class-direction}
\end{equation}
where $\cdscore(.)$ is the class direction score from~\cref{eqn:cfg}, $\mathrm{Downsample}$ and $\mathrm{Upsample}$ are downsampling and upsampling operations. We use a nearest-neighbors approach to prevent altering the statistics of the latent signal. In order to maintain the input to the diffusion models at the size $H \times W$, we dynamically pad the downsampled latent \(\xdown_t\) using a random background with a constant color. This encourages the model to concentrate content generation within the center area. We then crop the extended parts from the predicted noise to the target image resolution $N \times M$. Formally we modify the forward call for the reverse diffusion step as:
\begin{equation}
\begin{split}
    \xpad_t &\leftarrow \mathrm{Pad}\left(\xdown_t, \mathcal{A}_t\sqrt{\bar{\alpha}_{t}} + \sqrt{1 - \bar{\alpha}_{t}} \cdot \mathcal{Z}_t \right), \\
    \epsilon_{\theta}\left(\mathbf{x}_t, c\right) &= \mathrm{Crop} \left(\epsilon_{\theta}\left(\xpad_t, c\right), N \times M \right),
\end{split}
\label{eqn:pad-and-crop}
\end{equation}
where \(\mathcal{Z}_t \sim \mathcal{N}(0, I)\) represents the injected Gaussian noise at each step, and \(\mathcal{A}_t\)
represents a background image of size $(H - N) \times (W - M)$ with a constant color value 
$\pixel$ 
randomly sampled from a uniform distribution. 
This simple padding operation guarantees that the input to the diffusion model is kept at $H \times W$, while encouraging the diffusion model to keep the generated content within the cropped $N \times M$ center that we are interested in.
\subsection{Refined Class Direction Score}
\label{subsec:resampling}
\begin{figure}
    \centering
    \includegraphics[width=\linewidth]{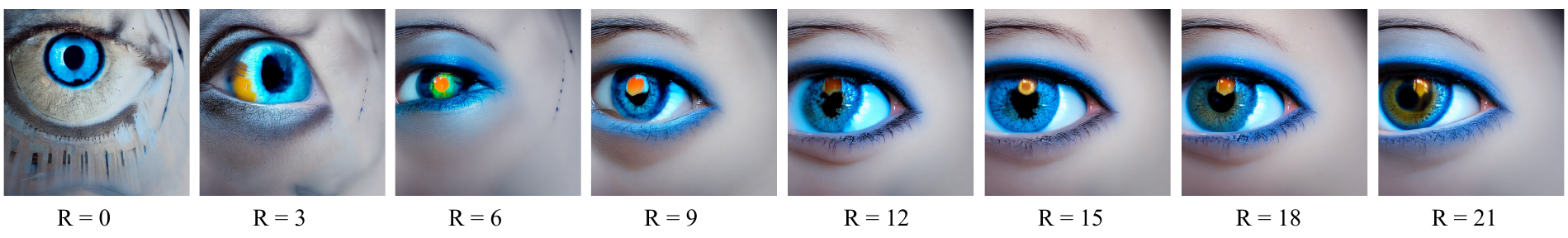}
    \vspace*{-0.3in}
    \caption{\textbf{Effect of Resampling.} Applying more resampling steps noticeably enhances detail in the generated images.}
    \label{fig:resampling}
    
\end{figure}

Sharing the class direction score among nearby pixels can result in over-smoothed images. To mitigate this, we propose an iterative resampling technique that increases the resolution of the estimated class direction score by extrapolating missing image components from their surrounding context, following ~\cite{repaint}. Our technique involves a gradual enhancement of the class direction score's resolution by estimating and integrating it for newly sampled pixels. Specifically, in each iteration, we replace $n\%$ of the pixels in $\xdown_t$ with newly sampled ones from $x_t$ to get an updated version $\tilde{\xdown}_t$. Following each update, the direction score is recalculated and blended with the previously calculated score. Formally, we consider $\mathbf{S}_d^0 = \mathbf{S}_d$ and define the iterative resampling as:
\begin{equation}
\label{eqn:resampling}
\begin{split}
\mathbf{S}_d^{r+1} = \mathbf{S}_d^r \odot m + \mathbf{\tilde{S}}_d \odot (1 - m),
\end{split}
\end{equation}
where $m \in \{0, 1\}^{\bar{H} \times \bar{W}}$ is a mask with a value of 1 at the positions of the newly sampled pixels and 0 elsewhere.  $\mathbf{\tilde{S}}_d$ represents the recalculated class direction score on $\tilde{\xdown}_t$ as per~\cref{eqn:class-direction}. This method estimates the class direction score of $n\%$ new pixels while retaining the information of the previously estimated score, thereby increasing the score's overall resolution. In our experiments, we set $n$ to $20\%$ and repeat the process $R$ times, depending on the target generation resolution.~\cref{fig:resampling} demonstrates the effectiveness of our method in enhancing the details of the generated images.
\begin{figure}[t!]
    \centering
    \includegraphics[width=\linewidth]{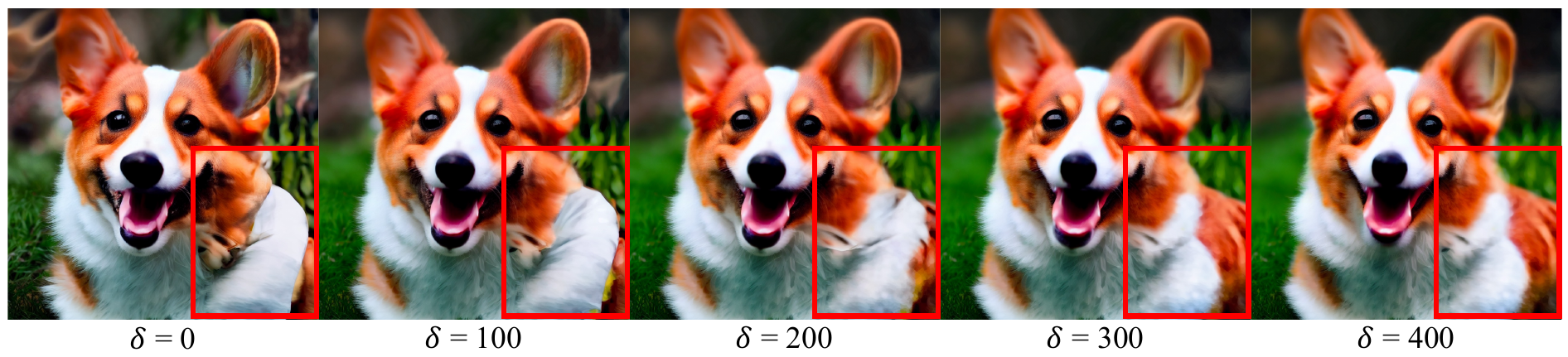}
    \vspace{-0.2in}
    \caption{\textbf{The Effect of Reduced-Resolution Guidance (RRG).} Higher RRG weights effectively eliminates emerging artifacts albeit at the cost of slightly blurrier outputs. $\delta = 200$ strikes a good balance. Improvements are more noticeable when zooming in.}
    \label{fig:rrg} 
    \vspace*{-0.1in}
\end{figure}

\begin{figure*}[t]
    \centering
    \includegraphics[width=0.96\textwidth]{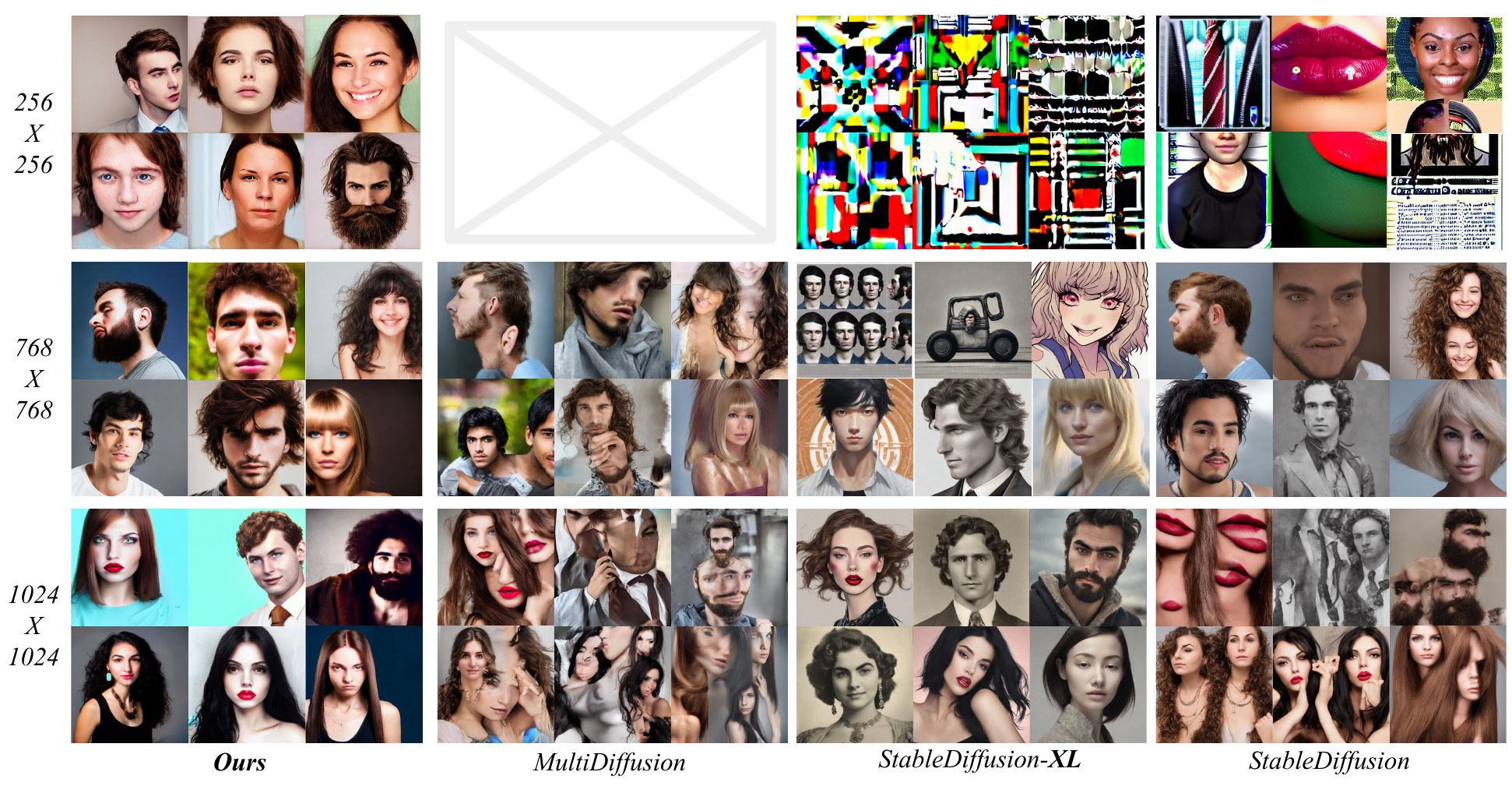}
    \vspace{-0.1in}
    \caption{\textbf{Qualitative comparison at various resolutions on CelebAHQ faces.} \emph{\name} consistently generates coherent images at all resolutions. \emph{StableDiffusion}, and \emph{MultiDiffusion} produce repeating body parts at higher resolutions, while \emph{StableDiffusion-XL} fails to maintain its quality at lower resolutions, resulting in noisy outputs at $256 \times 256$. We exclude \emph{MultiDiffusion} results at $256 \times 256$ as it is not designed to produce images at lower resolutions. }
    \label{fig:square-comparision}
\end{figure*}

\subsection{Reduced-Resolution Guidance (RRG)}
\label{subsec:rrg}
We effectively estimate the unconditional score signal and concurrently steer the image generation using the class direction score. However, inaccuracies in unconditional score estimation or fluctuations in local content, especially from distant patches, can lead to artifacts. To enhance image coherence and diminish artifacts, we consider a downsampled version of the latent at each decoding step as a reference and aim to align the decoded latent with it through our reduced-resolution latent update strategy. Specifically, we utilize the noise-free sample $\hat{x}_0^t$ of the latent $x_t$ from~\cref{eqn:ddim} and generate a corresponding noise-free sample $\hat{\xdown}_{0}^t$ from its downsampled counterpart $\xdown^t$ in~\cref{eqn:class-direction} as:{
\small
\begin{equation*}
\hat{\xdown
}^t_0 = \frac{1}{\sqrt{\bar{\alpha}_t}} (\xdown_t - \sqrt{1 - \bar{\alpha}_t} \cdot \left( \epsilon_{\theta}( \xdown_t) + (1 + w) \cdot \Delta_\mathcal{C}(\xdown_t, c) \right)),
\end{equation*}
}
Here, both $\hat{x}_0^t$ and $\hat{\xdown}^t_0$ corresponds to the same latent update at different resolutions.  Due to its smaller dimension, $\hat{\xdown}^t_0$ has a broader context when computing the unconditional signal.
To guide $x_t$ with its downsampled reference, we refine the latent $x_t$ with the direction that minimizes $L2$ difference between $\hat{\xdown}^t_0$  and $\hat{x}_0^t$. Formally, 
\begin{equation}
\label{eqn:rrg}
    \bar{x}_{t-1} \leftarrow \bar{x}_{t-1} - \delta_{t} \nabla_{x_{t}} ||\mathrm{Upsample}(\hat{\xdown}^t_0 , \bar{H} \times \bar{W}) - \hat{x}_{0}^{t}||,
\end{equation}
where $\delta_{t}$ represent the weight of the guidance. Since the overall image structure is determined in the early diffusion steps, we start with $\delta_{t} = 400$ and follow a cosine scheduler to decrease this weight for later diffusion steps. This scheduling strategy mitigates potential quality degradation from matching the generated image with a lower-resolution version while allowing the model to fill-in higher-resolution details in the later decoding stages.~\cref{fig:rrg} illustrates how RRG eliminates emerging artifacts.

\begin{figure}[t!]
    \centering
    \includegraphics[width=\linewidth]{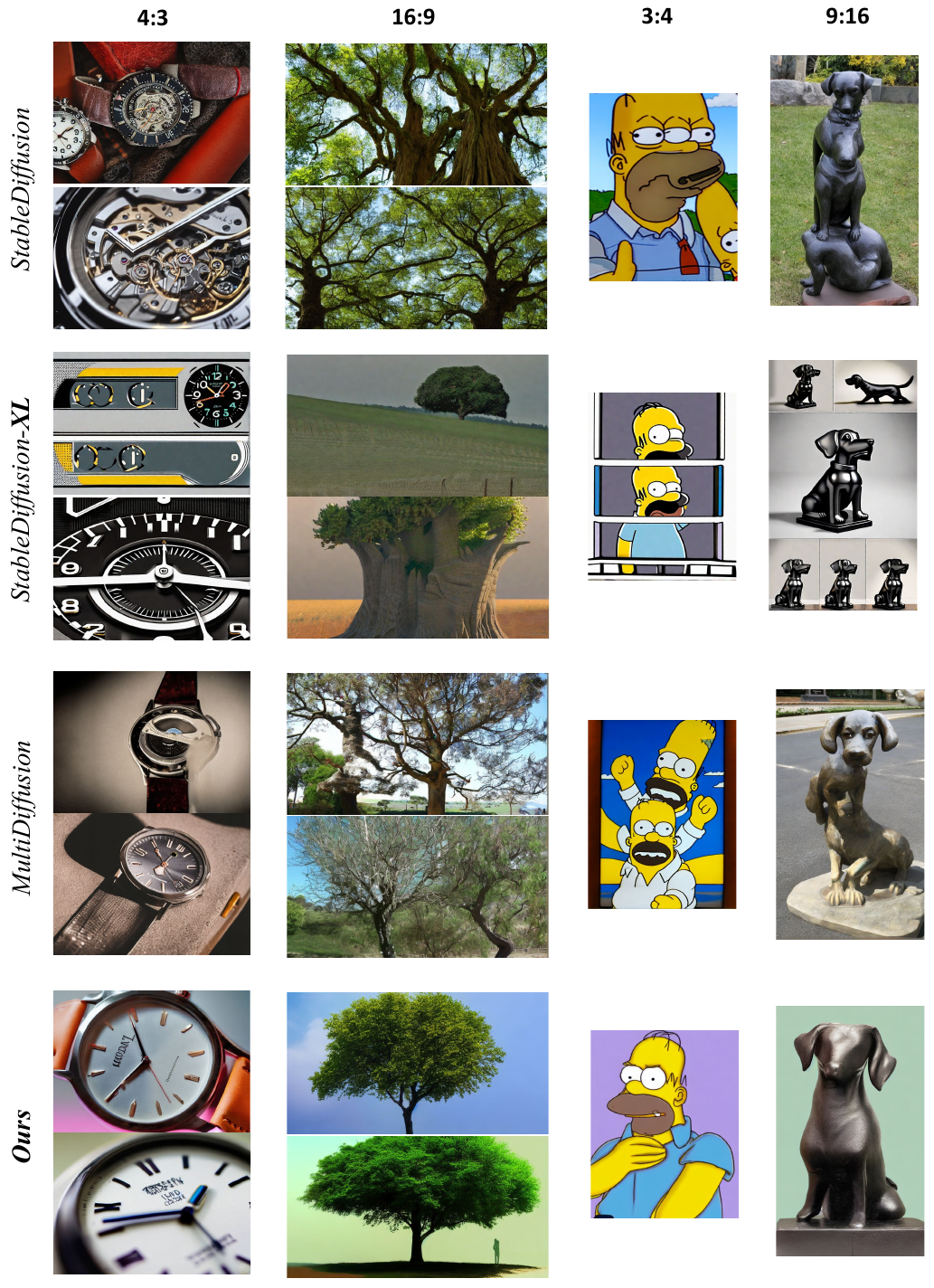}
    \caption{\textbf{Qualitative comparison on various aspect ratios.} \emph{\name} effectively handles a variety of aspect ratios. In comparison,  \emph{SD} and \emph{MultiDiffusion} generate unrealistic images, while \emph{SDXL} outputs exhibit a decline in the perceptual quality.}
    \label{fig:aspect-ratio-comparision}
    \vspace{-0.2in}
\end{figure}

\section{Experiments}
\label{results}
\emph{\name} supports generating images across different resolutions and aspect ratios. Our experiments focus on: (1) square images at multiple resolutions, and (2) images with varied aspect ratios and resolutions.

\noindent \textbf{Datasets.} We evaluate the generation of square images on the Multi-Modal \mbox{CelebAHQ} dataset~\cite{celeba}, which includes high-resolution square face images accompanied with text descriptions. We evaluate different aspect ratio generation using the LAION-COCO dataset~\cite{laioncoco}, derived from the web-crawled LAION-5B dataset \cite{laion5b}, which includes a variety of image aspect ratios and contains landscapes, people, objects, and everyday scenes, each paired with a synthetic caption generated using BLIP~\cite{blip}. We consider four common aspect ratios: 9:16, 16:9, 3:4 and 4:3.

\noindent \textbf{Evaluation Metrics.} Following prior text-to-image synthesis works~\cite{sdxl, imagen, unclip, multidiffusion, glide}, we use automatic evaluation metrics \emph{Frechet Inception Distance (FID)} and \emph{CLIP-score}. \emph{FID} \cite{fid} measures both the realism and diversity of the generated images by calculating the difference between features of the real and generated images computed using Inception-V3 ~\cite{inception-v3} pretrained on ImageNet \cite{imagenet}. \emph{CLIP-score} uses a pretrained text-image CLIP model~\cite{clip} to measure alignment between the generated images and input prompts. We use 10,000 images to compute these scores.

\noindent \textbf{Baselines.} 
We compare our approach against prior diffusion model generation methods, specifically focusing on \emph{Stable Diffusion (SD)} and \emph{MultiDiffusion (MD)}. \emph{SD} follows the standard reverse diffusion process on an image latent space. \emph{MD} uses a pretrained \emph{SD} for panoramic image generation, by creating smaller, overlapping patches and averaging the Diffusion Model scores in intersecting areas. For baseline comparisons, we fix the pre-trained diffusion model to $SD_{1.4}$, which is trained to generate images at a resolution of $512 \times 512$. Additionally, we compare our method with \emph{SDXL}, an enhanced SD model that is three times larger than the standard one. SDXL is trained at a higher resolution of $1024p$, and fine-tuned on a specific set of aspect ratios with pixel sums close to $1024^2$. We exclude the latent refinement module in \emph{SDXL} and focus our analysis on the base model. Throughout our experiments, we employ a DDIM sampling strategy with 50 steps and use $7.0$ for the classifier-free guidance scaling factor.  We also ensure consistency in seeds and captions for all baselines. 

\subsection{Qualitative Results}
We show square image generation samples in~\cref{fig:square-comparision} generated from the CelebAHQ dataset at various resolutions. Both \emph{MultiDiffusion} and \emph{Stable Diffusion} have a tendency to replicate textures and body parts at higher resolutions, resulting in images that lack coherence. At resolutions lower than its training resolution $512 \times 512$, \emph{StableDiffusion} aligns poorly with the provided captions and produces unappealing images. \emph{SDXL}, trained for $1024 \times 1024$ resolution, has a similar trend of reduced perceptual quality at lower sizes, eventually producing complete noise at a resolution of $256 \times 256$. In contrast, \name{} maintains image coherence across all tested resolutions, and yields results comparable to \emph{SDXL} at $1024\times1024$, despite using a less powerful base model and lower memory. We excluded results at $512 \times 512$ because both our method and \emph{MultiDiffusion} generate same outputs as the base \emph{Stable Diffusion} model. We also omitted the results of \emph{MultiDiffusion} at $256 \times 256$ since it is not designed to generate images at sizes smaller than the base model.
~\cref{fig:aspect-ratio-comparision} presents generated samples at various aspect ratios. We observe a similar trend of pattern repetition and reduced perceptual quality for images generated by the baselines, in contrast to those generated by \name{}. Finally, we apply our method using \emph{SDXL} as the base model to enable Full-HD image generation ($1920 \times 1080$) and provide examples in Supp.~\cref{supp:additional-results}.

\begin{table}[!t]
\centering
\caption{Quantitative comparison of on CelebA-HQ and LAION-COCO at different resolution. We indicate the best performances in \textbf{bold}, and \underline{underline} the second best ones. \emph{$\# Calls$} represent the number of diffusion model calls required at each decoding steps. }
\small
\setlength{\tabcolsep}{1pt}
\renewcommand{\arraystretch}{1.5}
\resizebox{\linewidth}{!}{

\begin{tabular}{cl c cc c cc c c c}

\toprule

\multirow{2}{*}{\bf \footnotesize Res.} & \multirow{2}{*}{\bf \footnotesize Method} && \multicolumn{2}{c}{\bf CelebA-HQ} && \multicolumn{2}{c}{\bf LAION-COCO} && \multirow{2}{*}{\bf \#Calls} & \multirow{2}{*}{\bf Mem.} \\

\cmidrule(lr){4-5} \cmidrule(lr){7-8}

&&& 
FID $\downarrow$ & CLIP $\uparrow$ && 
FID $\downarrow$ & CLIP $\uparrow$ &&& \\

\midrule

\multirow{3}{*}{\parbox{1cm}{\centering 256\\ $\times$\\ 256}}& 
$\text{SD}_{1.4}$ &&
\underline{258.43} & \underline{20.14} &&
\underline{54.06}  & \underline{21.43} &&
2 & 7.2GB \\

& SDXL  &&
368.06 & 14.40 &&
175.87 & 14.60 &&
2 & 18.5GB \\

& $\textbf{Ours}_{1.4}$ &&
\textbf{235.23} & \textbf{23.88} &&
\textbf{23.77} & \textbf{26.30} &&
2 & 8.6GB \\

\midrule

\multirow{2}{*}{\parbox{0.5cm}{\centering 512\\ $\times$\\ 512}} & 
$\text{SD}_{1.4}$  &&
233.40 & 24.00 &&
20.50 & 27.33 &&
2 & 8.6GB \\

& SDXL  &&
240.20 & 21.57 &&
42.58 & 25.34 &&
2 & 21.6GB \\

\midrule

\multirow{4}{*}{\parbox{1cm}{\centering 768\\ $\times$\\ 768}} &
$\text{SD}_{1.4}$  &&
238.87 & 23.45 &&
29.89 & 27.01 &&
2 & 11.1GB \\

& $\text{MD}_{1.4}$ &&
240.56 & 22.82 &&
29.98 & \underline{27.31} &&
50 & 8.6GB \\

& SDXL  &&
\textbf{225.48} & \underline{24.23} &&
\textbf{23.31} & \textbf{27.88} &&
2 & 24.5GB \\

& $\textbf{Ours}_{1.4}$ &&
\underline{225.86}  & \textbf{26.66} &&
\underline{25.78}  & 25.93 &&
17 & 8.6GB \\

\midrule

\multirow{4}{*}{\parbox{1cm}{\centering 1024\\ $\times$\\ 1024}}&
$\text{SD}_{1.4}$  &&
266.01 & 21.90 &&
47.01  & 25.70 &&
1 & 14.7GB \\

& $\text{MD}_{1.4}$  &&
264.57 & 21.55 &&
37.70 & \underline{26.96} &&
162 & 8.6GB \\

& SDXL  &&
\underline{230.21} & \textbf{24.62} &&
\textbf{25.58} & \textbf{28.06} &&
1 & 27.5GB \\

& $\textbf{Ours}_{1.4}$ &&
\textbf{228.87} & \underline{23.74} &&
\underline{27.76} & 26.07 &&
33 & 8.6GB \\

\bottomrule

\end{tabular}
}
\label{tab:square_comparison}
\vspace*{-0.2in}
\end{table}

\begin{table}[!t]
\centering
\caption{\textbf{Quantitative comparison of on LAION-COCO datasets at various aspect ratios (A) and resolutions (R).} best performances are in \textbf{bold}, andthe second best are \underline{underlined}. Vertical means the resolution is transposed from H:W to W:H.}
\small
\setlength{\tabcolsep}{4pt}
\renewcommand{\arraystretch}{1.2}
\resizebox{\linewidth}{!}{
\begin{tabular}{c  c l cc cc}

\toprule
\multirow{3}{*}{\bf A} 
& 
\multirow{3}{*}{\bf R} & \multirow{3}{*}{\bf Method} & \multicolumn{2}{c}{\bf Horizontal} & \multicolumn{2}{c}{\bf Vertical} \\
\cmidrule(lr){4-5} \cmidrule(lr){6-7}
& & & FID $\downarrow$ & CLIP$\uparrow$ & FID $\downarrow$ & CLIP$\uparrow$ \\
\midrule

\multirow{12}{*}{3:4} & 
\multirow{4}{*}{\parbox{1cm}{\centering 384\\ $\times$\\ 512}} & $\text{SD}_{1.4}$ & \underline{38.86} & \underline{24.63} & \underline{17.66} & \underline{26.54}  \\
&& $\text{MD}_{1.4}$ & -- & -- & -- & -- \\
&& SDXL & 104.84 & 21.33 & 68.84 & 22.40  \\
&& $\text{Ours}_{1.4}$ & \textbf{35.10} & \textbf{24.91} & \textbf{15.50} & \textbf{27.33}\\

\cmidrule{2-7}

&\multirow{4}{*}{\parbox{1cm}{\centering 512\\ $\times$\\ 680}} & $\text{SD}_{1.4}$ & 45.54 & \textbf{24.96} & \textbf{16.81} & \textbf{27.33} \\
&& $\text{MD}_{1.4}$ & \underline{43.44} & \underline{24.56} & 19.13 & 26.80  \\
&& SDXL & 62.80 & 24.20 & 28.23 & 26.17  \\
&& $\text{Ours}_{1.4}$ & \textbf{41.06} & 24.40 & \underline{18.90} & \underline{26.83}  \\

\cmidrule{2-7}

&\multirow{4}{*}{\parbox{1cm}{\centering 768\\ $\times$\\ 1024}} & $\text{SD}_{1.4}$ & 71.00 & 24.09 & 28.83 & 26.30 \\
&& $\text{MD}_{1.4}$ & 54.89 & \underline{25.02} & 26.35 & \underline{26.95} \\
&& SDXL & \underline{47.05} & \textbf{25.79} & \textbf{19.50} & \textbf{27.41} \\
&& $\text{Ours}_{1.4}$ & \textbf{47.03} & 24.91 & \underline{22.52} & 25.80 \\

\midrule

\multirow{8}{*}{9:16} & 
\multirow{4}{*}{\parbox{1cm}{\centering 288\\ $\times$\\ 512}} &
 $\text{SD}_{1.4}$ & \underline{23.50} & \underline{24.69} & \underline{24.01} & \underline{24.89} \\
&& $\text{MD}_{1.4}$ & -- & -- & -- & -- \\
&& SDXL & 121.83 & 17.65 & 112.41 & 18.54 \\
&& $\text{Ours}_{1.4}$ & \textbf{23.23} & \textbf{25.26} & \textbf{22.86} & \textbf{26.30} \\

\cmidrule{2-7}

&\multirow{4}{*}{\parbox{1cm}{\centering 512\\ $\times$\\ 904}} & $\text{SD}_{1.4}$ & 29.86  & 25.34 & 27.45 & 26.01 \\
&& $\text{MD}_{1.4}$ & \underline{26.35} & \textbf{25.70} & \underline{26.70} & 25.28 \\
&& SDXL & 29.60 & \underline{25.40} & 27.27 & \underline{26.08} \\
&& $\text{Ours}_{1.4}$ & \textbf{22.85} & 25.01 & \textbf{26.68} & \textbf{26.12} \\

\bottomrule
\end{tabular}
}
\label{tab:combined_comparison}
\vspace*{-0.1in}
\end{table}

\subsection{Quantitative Results} 
Table~\ref{tab:square_comparison} shows quantitative evaluations. \emph{StableDiffusion} shows increasing image quality degradation, as indicated by the \emph{FID} metric when processing images of sizes different from its training resolution $512 \times 512$. \emph{MultiDiffusion} slightly improves \emph{FID} at the expense of a substantially more forward calls. \emph{SDXL} demonstrates similar declines in quality at resolutions far from its fine-tuning size of $1024 \times 1024$. \name{}, however, improves \emph{FID} while maintaining a comparable \emph{CLIP-score} to the base model. \name{} also significantly improves the performance for lower resolutions $256 \times 256$, while achieving comparable results to \emph{SDXL} at its training resolution $1024 \times 1024$, with only $\sim 31\%$ of the memory requirement for \emph{SDXL}. 

~\Cref{tab:combined_comparison} provides an evaluation on a variety of aspect ratios and resolutions for both %
horizontal and vertical image resolutions. \emph{StableDiffusion} obtains worse \emph{FID} score with increasing resolutions, while surprisingly maintaining or improving its \emph{CLIP-score}. We posit that since \emph{CLIP-score} quantifies the agreement between the input prompt and the generated image, \emph{StableDiffusion} benefits from generating repetitive textures and artifacts that align closely with the prompt. For example, a photo of repeated lipsticks might align more with the caption \emph{"lipstick"} despite its poor composition. \emph{MultiDiffusion} generally enhances image quality but does not achieve satisfactory performance. \emph{MultiDiffusion} struggles with objects that span the entire image (\eg~\cref{fig:aspect-ratio-comparision}). Our method consistently improves \emph{FID} over the baseline on horizontal and most vertical resolutions while preserving fidelity to the input prompts, thereby attaining comparable or superior \emph{CLIP-scores}. Remarkably, even with a larger model size and explicit fine-tuning at a similar resolution of $768 \times 1280$, \emph{SDXL} only marginally surpasses our method at the $768 \times 1024$ resolution. 
\subsection{Ablation study} 
\begin{table}[!t]
\centering
\caption{\textbf{Ablation analysis of \name} on 500 images.}
\setlength{\tabcolsep}{4pt} 
\resizebox{0.84\columnwidth}{!}{
\noindent\begin{tabular}{lccc}
    \toprule
\hspace{-2mm}\bf Model Details &FID $\downarrow$ & CLIP $\uparrow$  \\
    \midrule
\hspace{-1mm}\name & 133.67  & 25.82  \\
\hspace{-1mm}$\quad$ w/o \emph{Resampling} & 150.01  & 23.82 \\
\hspace{-1mm}$\quad$ w/o \emph{RRG} & 150.64 & 24.34 \\
\hspace{-1mm}$\quad$ w/o \emph{Imp. overlap}, w/ exp. overlap & 141.42 & 25.82 \\
    \bottomrule
\end{tabular}
}
\label{tab:ablation_table}
\vspace{-4mm}
\end{table}

\cref{tab:ablation_table} presents results of our method when excluding key components, demonstrating that each element improves performance. \cref{fig:context-pixs} illustrates the effectiveness of our proposed implicit overlap method in resolving boundary discontinuities at a reduced computational cost. \cref{fig:rrg} highlights the effectiveness of \emph{Reduced-Resolution Guidance} in removing emerging artifacts. \cref{fig:resampling} shows the effectiveness of our iterative resampling technique in enhancing details.

\section{Discussion and Conclusion}
\label{sec:conclusion}
Experimental results highlight the adaptability and effectiveness of \name{} at steering diffusion models to produce an array of resolutions and aspect ratios. \name{} requires no fine-tuning, consumes a constant memory footprint, enables both increased and reduced resolutions, and can generate a variety of aspect ratios. 

\name{}, however, does have several practical limitations. First, inaccuracies in estimating the global and local signals may result in artifacts. Although we attempt to mitigate artifacts with our \emph{Reduced-resolution guidance}, it can still generate blurrier outputs. Second, since the global content guidance is initially estimated at the original training resolution of the underlying diffusion model, our method is less effective in generating images at significantly \emph{extended} sizes beyond the training resolution. In particular, at extreme resolutions beyond 4X, our method produces artifacts and images of a lower perceptual quality. We provide examples of these failure cases in Supp.~Sec.~10.4.

The main insight underpinning our method is a novel reinterpretation of \emph{classifier-free guidance} in somewhat disentangling both global and local content. Our comprehensive evaluations demonstrate the feasibility of disentangling these signals, yet the full extent of their separation offers an avenue for further exploration in this direction.  We hope that our findings inspire future work in investigating the separation of global and local content guidance signals for image synthesis. This separation holds potential for various applications such as selectively manipulating local and global content or enhancing style transfer. Additionally, the rich semantic representation in the class-direction score has the potential for improving discriminative models.

\vspace{0.02in}
\noindent
\textbf{Acknowledgments.} This work was partially funded by NSF Award \#2201710.

{
    \small
    \bibliographystyle{ieeenat_fullname}
    \bibliography{main}
}

{\onecolumn
\section*{Appendices}
In this supplementary document, we extended the discussion on existing diffusion models generation processes, highlighting their constraints in adapting to diverse image sizes and the potential for separating global and local content generation. We also provide further qualitative comparisons with baselines using the DrawBench benchmark \cite{glide} at various resolutions including full-HD. Our code can be accessed at 
\textcolor{blue}{\href{https://github.com/MoayedHajiAli/ElasticDiffusion-official.git}{https://github.com/MoayedHajiAli/ElasticDiffusion-official.git}}

\section{\name Symbols and Implementation.}
To simplify the notation in this paper, we have employed specific symbols to denote the key elements within our framework. To clarify the associations between each symbol and their meanings, we provide in Tab.~\ref{tab:symbols}, a detailed explanation of each symbol that we used in the paper. Additionally, we describe in \cref{alg:sampling} the full generation process of an image or arbitrary size $\bar{H}\times \bar{W}$ using a pre-trained base diffusion models that operates on images of size $H\times W$
\begin{table}[H]
\centering
\caption{Table of symbols used in this paper.}
\label{tab:symbols}
\begin{tabular}{cl}
\toprule
\textbf{Symbol} & \textbf{Correspondence} \\
\midrule
\( H\times W \) & training resolution of the base diffusion model \\
\( \bar{H}\times \bar{W} \) & target resolution of the generated image \\
\( N \times M \) & chosen resolution of the same aspect ratio as $\bar{H}\times \bar{W}$ but smaller than \( H\times W \) \\
\({\epsilon}_{\theta} \) & pre-trained diffusion model network\\
\(w \) & classifier-free guidance weight\\
\( x_t \) & diffusion latent at timestep $t$ of size $H\times W$\\
\( \bar{x}_t \) & diffusion latent at timestep $t$ of size $\bar{H}\times \bar{W}$\\
\( \xdown_t \) & downsampled latent from $\bar{x}_t$ \\
\( \hat{\xdown}_t \) & padded $\xdown_t$ to match the training resolution $H\times W$\\
\( \hat{x}_0^t \) & a noise-free sample of $x$\\
\( \hat{\xdown}_0^t \) & a noise-free sample of $\xdown$\\
\( p_k \) & a crop of $\bar{x}_t$ smaller than the training resolution\\
\( c_k \) & a context crop of $\bar{x}_t$\\
\( \mathbf{S}_u \) & unconditional score for latent at size $H \times W$\\
\( \mathbf{S}_c \) & conditional score for latent at size $H \times W$ \\
\( \mathbf{S}_d \) & class-direction score for latent at size $H \times W$\\
\( \bar{\mathbf{S}}_u \) & unconditional score for latent at size $\bar{H} \times \bar{W}$\\
\( \bar{\mathbf{S}}_d \) & class-direction score for latent at size $\bar{H} \times \bar{W}$\\

\bottomrule
\end{tabular}
\end{table}

\begin{algorithm}
\caption{Sampling Algorithm for Image at $\bar{H} \times \bar{W}$}
\begin{algorithmic}[1]
\Require 
\Statex $\dm$ \Comment{pre-trained DM at $H \times W$}
\Statex $c, w$ \Comment{text condition and CFG weight}
\Statex $\bar{x}_T \sim \mathcal{N}(0, I)$ \Comment{noise at $\bar{H} \times \bar{W}$}
\For{$t = T$ \textbf{down to} $1$}
    \State $\mathbf{x}_t \gets \mathrm{Downsample}(\bar{x}_t, N \times M)$
    \State $\mathcal{Z}_t \sim \mathcal{N}(0, I)$ 
    \State $\mathcal{A}_t \gets \pixel_{H-N, W-M}, \pixel \sim \text{Uniform}(0, 255)$ 
    \State $\mathbf{\hat{x}}_t \gets \mathrm{Pad}\left(\mathbf{x}_t, \mathcal{A}_t \sqrt{\bar{\alpha}_{t}} + \sqrt{1 - \bar{\alpha}_{t}} \cdot \mathcal{Z}_t \right)$ \Comment{Pad to match training resolution $H \times W$}
    \State $\mathbf{S}_c \gets \mathrm{Crop} \left(\epsilon_{\theta}\left(\hat{\mathbf{x}}_t, c\right), N \times M \right)$ \Comment{conditional score at target aspect ratio}
    \State $\mathbf{S}_u\leftarrow   \tilde{\epsilon}_{\theta}(\hat{\mathbf{x}}_t)$  \Comment{Uncodnitional score from Eq. (3)}
    \State $\mathbf{\bar{S}}_d^0 \gets \mathrm{Upsample}(\mathbf{S_c} - \mathbf{\hat{S}_u}, \bar{H} \times \bar{W})$ \Comment{class-direction score}
    
    \ForAll{$r = 1, \dots , R$} 
        \State $\mathbf{\bar{S}}_d^r \gets \mathrm{Resample}(\mathbf{\bar{S}}_d^{r-1}, \bar{x}_t)$ \Comment{Eq. (6)}
    \EndFor
    \State $\mathbf{\bar{S}_u} \Leftarrow \tilde{\epsilon}_{\theta}(\bar{x}_t)$  \Comment{Eq. (3)}
    \State $\bar{x}_{t-1} \gets \mathbf{\bar{S}}_u+ (1 + w) \cdot \mathbf{\bar{S}}_d^R$ \Comment{diffusion update}
    \State $\bar{x}_{t-1} \gets \bar{x}_{t-1} - \mathrm{RRG}(\bar{x}_t, \mathbf{x}_t)$ \Comment{Eq. (7)}
\EndFor
\State \textbf{return} $\bar{x}_0$
\end{algorithmic}
\label{alg:sampling}
\end{algorithm}

\section{Discussion on Diffusion Models}
In this section, we discuss the generation process of diffusion models, focusing on their performance across various image sizes and our analysis of their capacity to separate global and local content generation.
\subsection{Diffusion Models Adaptability Across Sizes}
Pretrained diffusion models such as $\textit{StableDiffusion}_{1.4}$ are technically capable of handling various image sizes. Accordingly, the official implementation provides parameters for adjusting the size of the generated images. However, our experiments show a significant decline in image quality when these models operate at resolutions outside those seen during training. These observations are confirmed in the Stable Diffusion blog post on Hugging Face which warns that deviating from the trained resolution may compromise image quality~\cite{sd-blog}. Specifically, it notes that going below the training resolution results in lower quality images, while exceeding it in both the height and width directions causes repetitive image areas, leading to a loss of global coherence. Similar findings were noted in the \emph{StableDiffusion-XL} official blog post~\cite{sdxl-blog}.

In Fig.~\ref{fig:supp-gradual-degredation}, we qualitatively analyze
how generating images larger than the training resolution impacts image coherence. We generate these results using $\textit{StableDiffusion}_{1.4}$ which was pretrained on $512 \times 512$ images. For smaller dimensions, the model tends to stretch the generated objects, whereas for larger dimensions, such as $1024 \times 1024$, it often creates repetitive elements. Notice the stretch in the cat and lion faces in the third and fourth columns. Additionally, observe how artifacts and repetition regarding nose and eye parts tend to happen more frequently as we increase the resolution.

Notably, the model maintains its output quality within a narrow margin of $64$ pixels from its training resolution, suggesting a limit to the generalization capabilities of diffusion models with respect to various image sizes. This observation also shows the potential limitations of the solutions based only on a fine-tuning process for a fixed set of aspect ratios such as those proposed in prior work%
~\cite{sdxl, any-size-diffusion}.%
\subsection{Global and Local Content Generation}
\label{global-local-analysis}
In the domain of generative adversarial networks (GANs), the disentanglement style and content in the synthesized images has been widely explored, paving the way for advancements in diverse generation and editing applications \cite{singan, vidstyleode, stylegan2}. However, the precise definitions of 'style' and 'content' remain fluid, with no consensus on the definition in the literature. Previous works often define the content and style based on manually pre-defined attributes. In this work, we opt to avoid such ambiguity by denoting the overall composition of the image as global content and the fine-grained details as local content. Subsequently, we conceive \name based on two key insights: First, the \emph{class direction score} (Eq. (2) in the main paper) collectively influences pixels to shape the overall composition of the generated image, denoted as global content. This global score can be effectively shared among neighboring pixels. Fig. ~\ref{fig:supp-sharing-class-direction-score} demonstrates that sharing the \emph{class direction score} between nearby pixels maintains the global content and coherence of the generated image, although increasing the sharing extent decreases the perceptual quality. In contrast, the  \emph{unconditional score} requires pixel-level precision and it may not be feasible to share it between nearby pixels, as illustrated in Fig.~\ref{fig:supp-sharing-uncond-score}. Second, the \emph{unconditional score} dictates the fine-grained details of the generated image, denoted as local content. This suggests that the score can be computed effectively on localized regions without necessitating global information from the entire image. Fig.~\ref{fig:supp-localized-scoring} shows that computing the \emph{unconditional score} in localized regions, corresponding to the size $512 \times 512$ of the generated image, produces similar results to those obtained when computing the score globally.

\section{Analysis of \name}
This section provides a comprehensive analysis of \name, focusing on its application to pixel-based diffusion models and comparisons with baseline methods. We present further qualitative results to showcase \name's effectiveness in enhancing the coherence of the generated image across various sizes. We finally discuss the limitations and failure cases of our method. 

\subsection{Additional Ablation Study.} To better understand the effect of the class-direction refinement and reduced-resolution guidance (RRG) strategy on the overall quality of the results, we analysed in Tab.~\ref{tab:rebuttal-ablation} the performance of \name when excluding these components at two different resolutions. We observe that their necessity is pronounced at higher resolutions (e.g.,~4x), while their influence is limited at lower upsampling scale (e.g.,~2x). Notably, even without these components, our method achieves \emph{better} FID than baselines.
\begin{table}[ht]
\setlength{\tabcolsep}{1mm} 
\centering
\noindent
\renewcommand{\arraystretch}{0.6} %
\begin{tabular}{@{\hspace{0.2cm}}l@{\hspace{0.2cm}}c@{\hspace{0.2cm}}c@{\hspace{0.2cm}}c@{\hspace{0.2cm}}c@{\hspace{0.2cm}}}
\toprule
& \multicolumn{2}{c}{$768\times768$} & \multicolumn{2}{c}{$1024\times1024$} \\
\cmidrule(lr){2-3} \cmidrule(lr){4-5}
& FID $\downarrow$ & CLIP $\uparrow$ & FID $\downarrow$ & CLIP $\uparrow$ \\
\midrule
Ours & 225.86 & 26.66 & 228.87 & 23.74 \\
\thinspace w/o RRG & 230.27 & 24.13 & 234.66 & 23.15 \\
\thinspace w/o RGG \& refined class-direction score & 233.49 & 23.32 & 263.15 & 20.91 \\
\bottomrule
\end{tabular}
\vspace{-0.1in}
\caption{\textbf{Ablation study} on CelebAHQ Dataset using $\textit{StableDiffison}_{1.4}$ as the base model.}
\label{tab:rebuttal-ablation}
\end{table}
\subsection{Generalization to Pixel-Based Diffusion Models}
We apply \name to a pre-trained \emph{DeepFloyd-IF-XL-V1.0} model, which operates on the pixel-space~\cite{deepfloyed}. In the first stage, \emph{DeepFloyd-XL-V1.0} generates images at a $64 \times 64$ resolution, which are then upscaled to $256 \times 256$ and subsequently to $1024 \times 1024$ in later stages. To assess the generalization capabilities of our method, we only focus on the first stage which generates images at a low resolution. As illustrated in Fig.~\ref{fig:supp-deepfloyed}, \emph{DeepFloyd-XL-V1.0} shows similar limitations as latent diffusion models when dealing with various resolutions, primarily characterized by repetitive elements and reduced image coherence. However, we demonstrate the effectiveness of \name in enhancing the ability of the pre-trained model to handle diverse resolutions and aspect ratios by successfully generating well-structured images. This shows the applicability of our proposed generation process to diffusion models that operate on the pixel space. 

\subsection{Additional Qualitative Results}
\label{supp:additional-results}
We provide further qualitative analysis of \name.

Figure~\ref{fig:supp-landscape-comparison} provides a comparison with \emph{StableDiffusion} and \emph{MultiDiffusion} using selected DrawBench \cite{imagen} prompts for horizontal images at resolution $680 \times 512$. We highlight the tendency of baseline methods to generate repetitive elements. This not only disrupts the image's overall coherence but also makes the baselines struggle to accurately reflect object counts. For example, in the first row, both baselines produced multiple dogs for an input prompt '\emph{one} dog on the street'. In contrast, our method effectively aligns with the given prompt, generating a single, coherent dog.

Figure~\ref{fig:supp-portrait-comparison} shows a similar analysis on vertical images of resolution $512 \times 680$. We observe a similar limitation in baselines such as element and texture repetition in the generated images. This tendency of repeating elements particularly affects the model's capacity to create coherent objects that share textures with the background. For example, In the first row, the baselines struggle to accurately depict a hamburger, whereas our method successfully generates a coherent hamburger that is separated from the background. This limitation also affects the baseline models' ability to render objects with repeating patterns, like a 'cube made of bricks' shown in the last row. Moreover, the baseline behavior of repeating patterns especially escalates when generating a single object across the majority of the image, as observed in the $4^{th}$ and $5^{th}$ rows. In contrast, our method is able to maintain image coherence across various settings.

Figures~\ref{fig:supp-landscape-sdxl-comparison-1} and~\ref{fig:supp-landscape-sdxl-comparison-2} focus on showing results for the generation of Full-HD horizontal images using \emph{StableDiffusion-XL} as the base model. We compare against the standard decoding process of \emph{StableDiffusion-XL} on sampled DrawBench prompts from the Reddit Category and observe a significant improvement in image coherence when applying our method. Notice in Fig.~\ref{fig:supp-landscape-sdxl-comparison-1} how \emph{StableDiffusion-XL} stretch the car in the first example, or repeat the limbs of the corgi and the lion (in the $2^{nd}$ and $3^{rd}$ example). In comparison, our method successfully coherently generates the requested image without any such distortions, all while utilizing lower memory requirements.

Figures~\ref{fig:supp-portrait-sdxl-comparison-1} and~\ref{fig:supp-portrait-sdxl-comparison-2} provide a similar analysis for Full-HD vertical images. \emph{StableDiffusion-XL} produces significantly distorted images which either contain incorrectly repeating elements as seen in the cat and the man faces in the first two examples of Fig.~\ref{fig:supp-portrait-sdxl-comparison-1}, or stretched parts like the woman face in the first example of Fig. ~\ref{fig:supp-portrait-sdxl-comparison-2}. In contrast, our method generates detailed objects that fit the vertical aspect ratio while avoiding any stretching or element repetition.

\subsection{Limitations}
Fig.~\ref{fig:supp-limitations} illustrates the limitations of \name in various scenarios. Our method retains some limitations of the pre-trained base model, including challenges with text-image alignment for complex prompts and occasional occurrence of artifacts. Additionally, we observe an increase in image blurriness with the application of larger \emph{Reduced-Resolution Guidance} weights (Sec. 4.4 of the main paper). Moreover, while infrequent, there are instances where the constant-color background inadvertently blends into the generated image (as discussed in Sec. 4.2 of the main paper).

\begin{figure*}
    \centering
    \includegraphics[width=\textwidth]{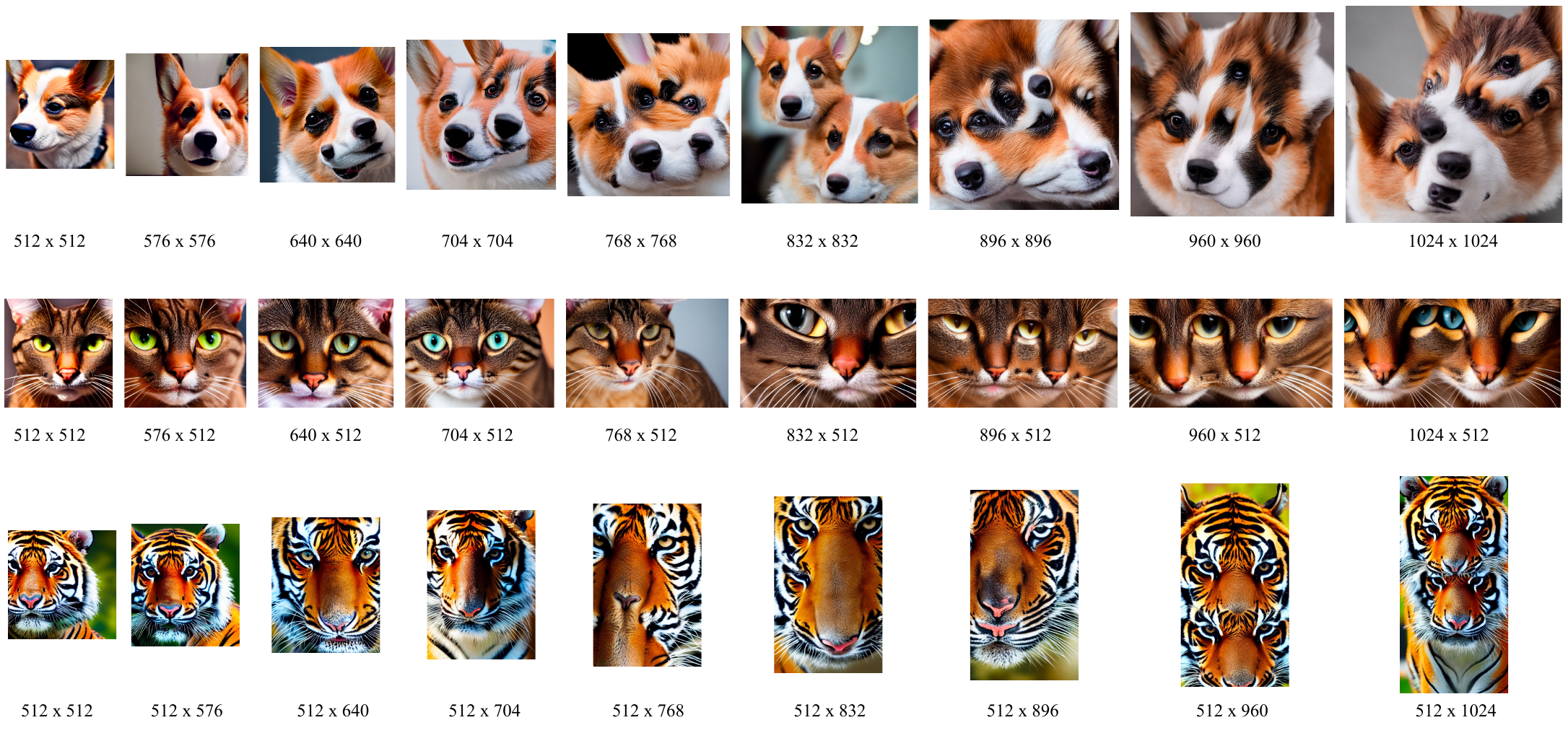}
    \caption{\textbf{Degradation of image quality in $\textbf{StableDiffusion}_{\textbf{1.4}}$ with varying resolutions.} We illustrate the progressive decrease in image quality as the resolution deviates from the model's training size of $512 \times 512$. The results on square, horizontal, and vertical resolutions show a significant reduction in global coherence for image sizes beyond 64 pixels from the training resolution.}
    \label{fig:supp-gradual-degredation}
\end{figure*}

\begin{figure*}
    \centering
    \includegraphics[width=\textwidth]{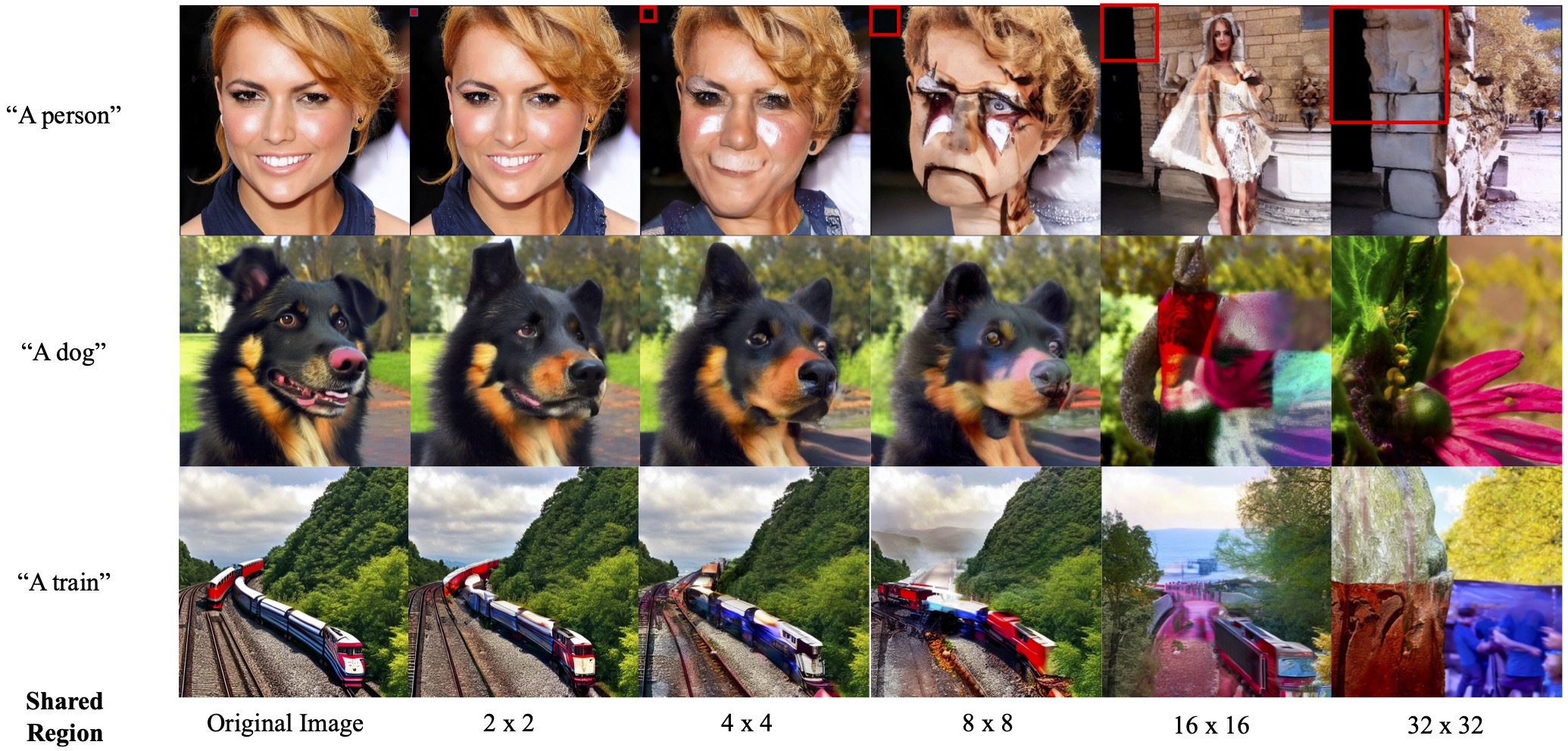}
    \caption{\textbf{Effect of sharing \textit{class direction score} between nearby pixels.} We highlight that sharing the score within a group of neighboring pixels preserves the global content and coherence of the image, despite a reduction in the perceptual quality when selecting larger group sizes (as denoted by the red square). This supports our assumption that this score tends to be similar among neighboring pixels. To conduct this experiment, we downsample the \emph{class direction score} of size $64 \times 64$ by a factor $N \times M$ (as specified in the last row) and upsample it back to $64 \times 64$, thus sharing the score for each $N \times M$ region. Note that as our experiments utilize a latent diffusion model, sharing the score within an $N \times M$ latent pixels during the decoding process impacts $8N \times 8M$ pixels of the final generated image.}
    \label{fig:supp-sharing-class-direction-score}
\end{figure*}

\begin{figure*}[htbp]
    \centering
    \includegraphics[width=\textwidth]{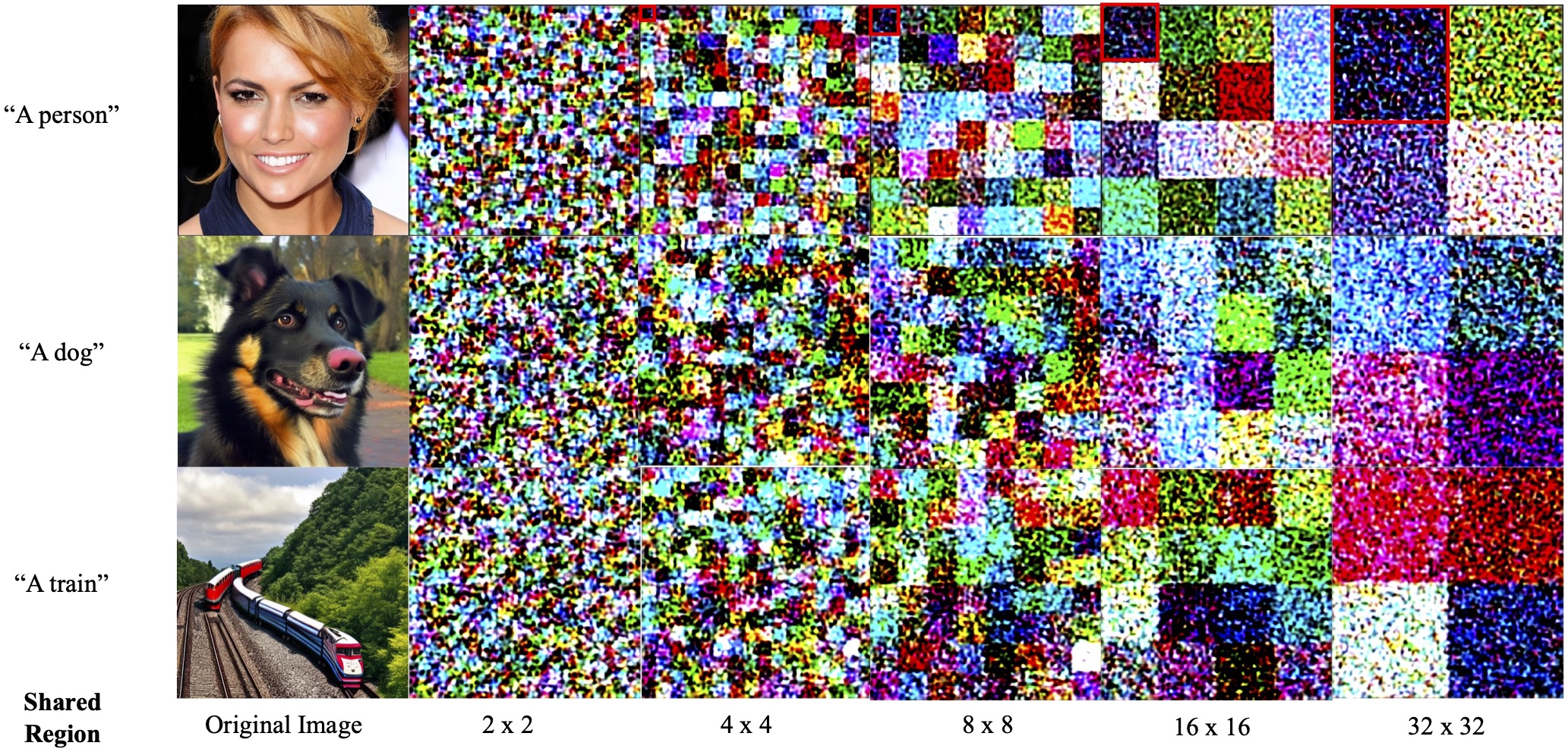}
    \caption{\textbf{Effect of sharing \textit{unconditional score} between nearby pixels.} We show that sharing the unconditional score, even in small groups of pixels, leads to the generation of complete noise. This indicates that \emph{uncondtional score}, in contrast to the \emph{class direction score}, requires pixel-level precision to generate local details.}
    \label{fig:supp-sharing-uncond-score}
\end{figure*}

\begin{figure*}
    \centering
    \includegraphics[width=\textwidth]{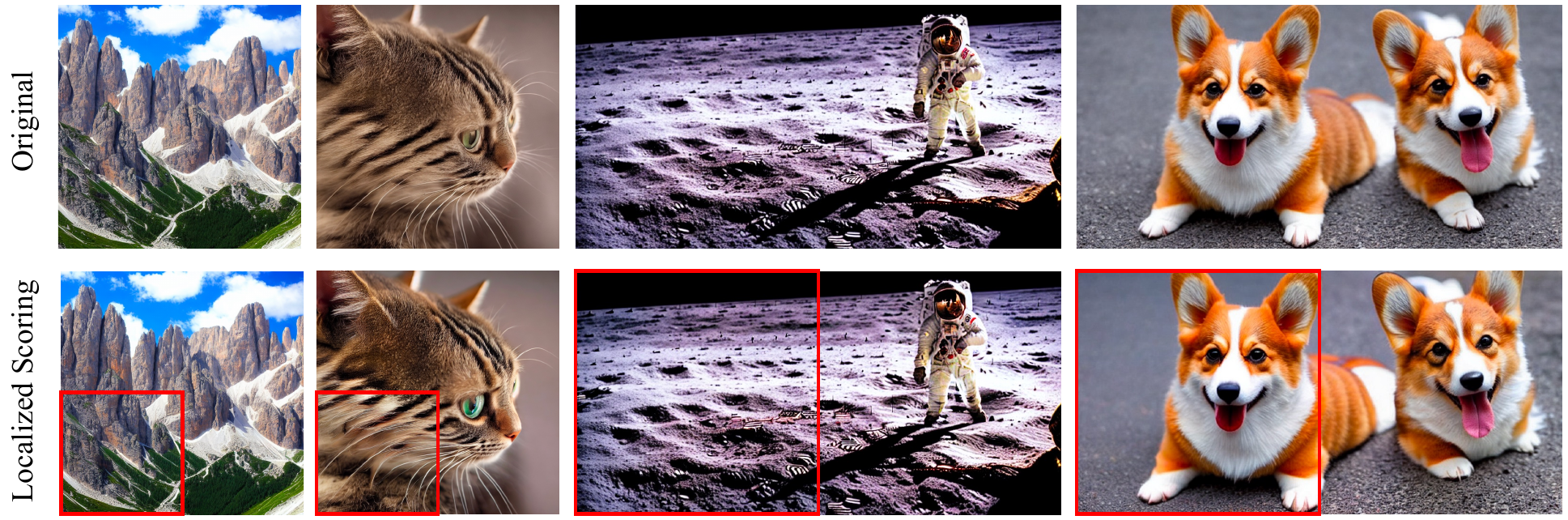}
    \caption{\textbf{Unconditional score computation on localized regions.} We show that computing the diffusion model \emph{unconditional score} on local patches  (corresponding to the size of the red boxes in the second row) results in images that are visually similar to those produced by a global score computation (displayed in the first row).}
    \label{fig:supp-localized-scoring}
\end{figure*}

\begin{figure*}
    \centering
    \includegraphics[width=\textwidth]{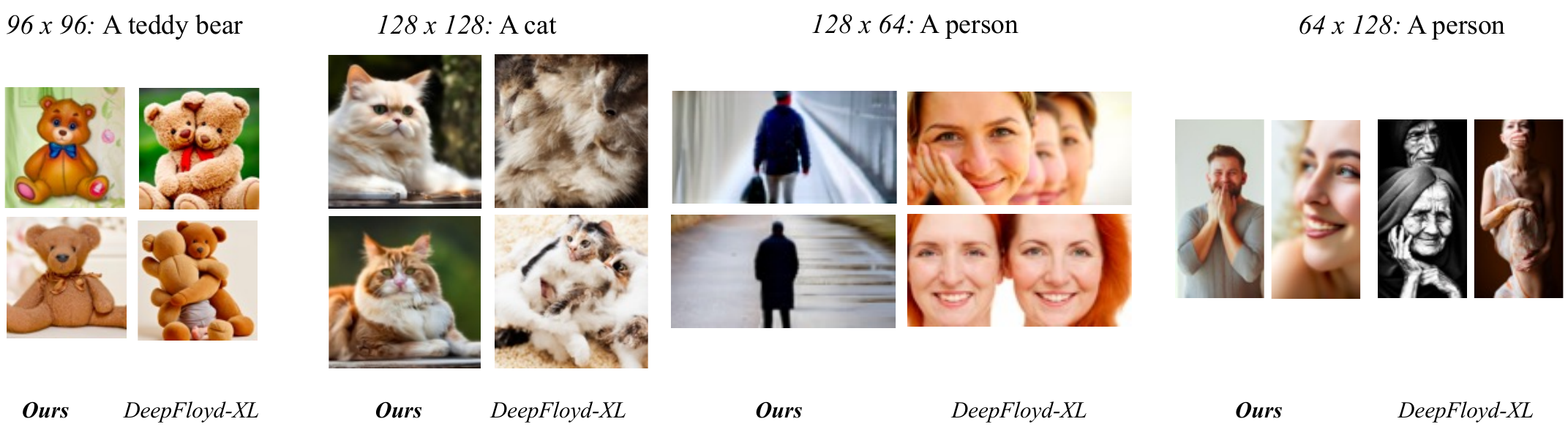}
    \caption{\textbf{\textit{DeepFloyd-XL} across various image sizes.} We test \textit{DeepFloyd-XL}, a diffusion model that operates on the pixel space, across multiple image resolutions. We observe a degradation in performance similar to that seen in $StableDiffusion$. The application of \name significantly improves the overall composition of the generated images.}
    \label{fig:supp-deepfloyed}
\end{figure*}

\begin{figure*}
    \centering
    \includegraphics[width=0.95\textwidth]{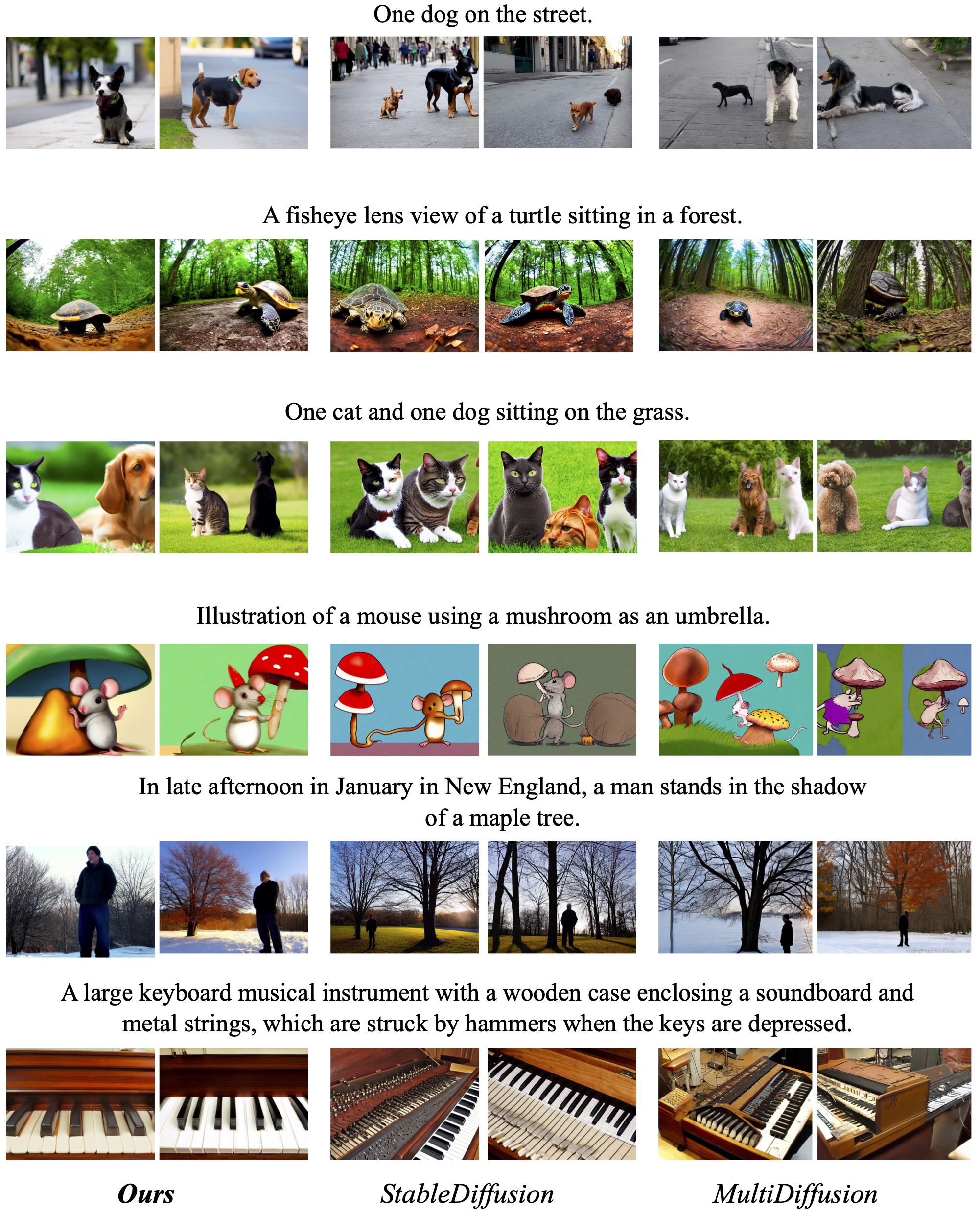}
    \caption{\textbf{Additional qualitative comparison on horizontal images using selected DrawBench prompts.} We use $SD_{1.4}$ as a base model for our method, \textit{StableDiffusion}, and \textit{MultiDiffusion} and generate images at resolution $680 \times 512$. Images produced by baselines display reduced alignment to the input prompt ($1^{st}$ and $3^{rd}$ rows) and repeated elements ($4^{th}$ and $6^{th}$ rows). In comparison, our method displays superior image coherence and faithfulness to the input prompts.}
    \label{fig:supp-landscape-comparison}
\end{figure*}

\begin{figure*}
    \centering
    \includegraphics[width=0.85\textwidth]{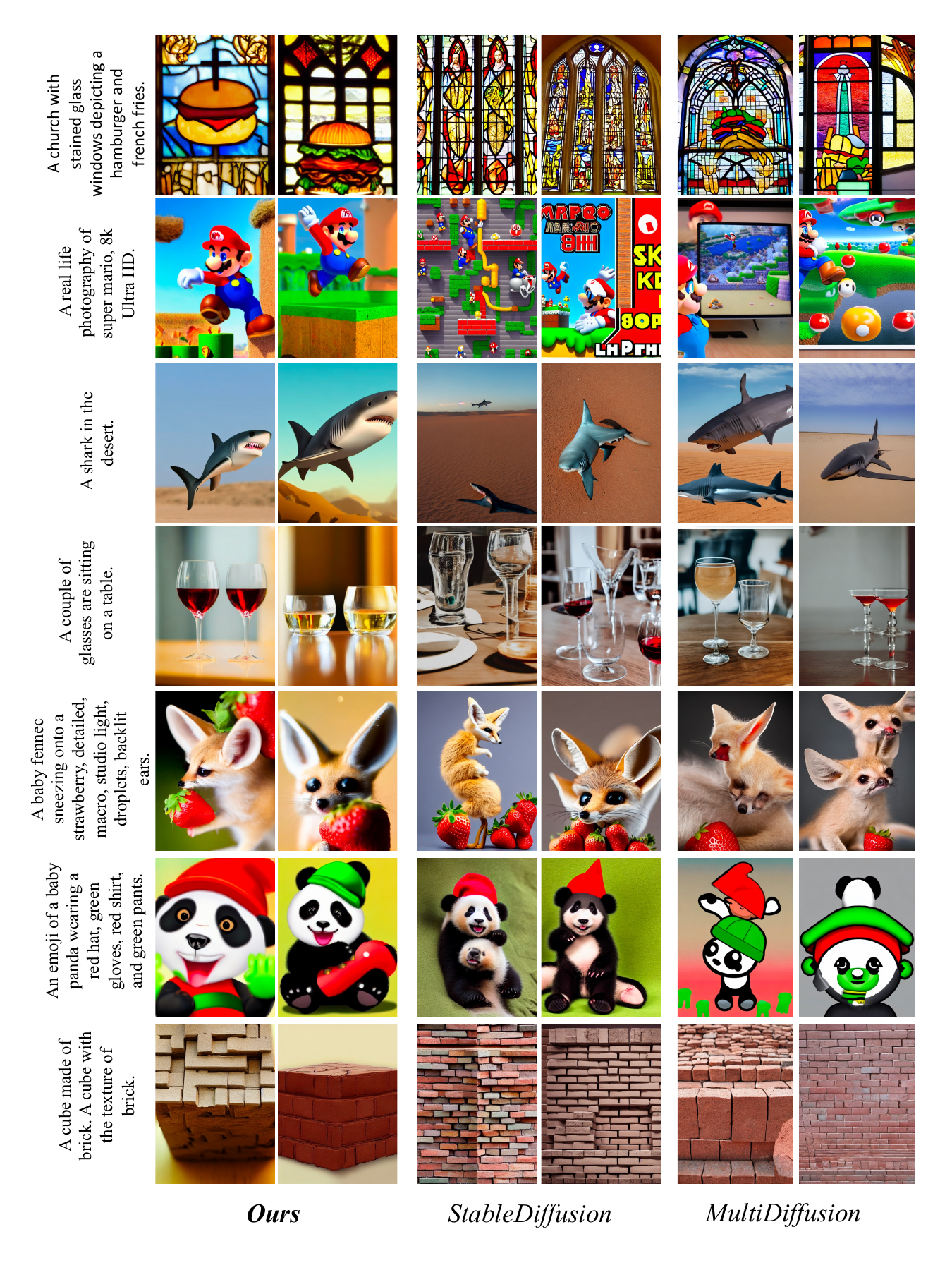}
    \caption{\textbf{Additional qualitative comparison on vertical images using selected DrawBench prompts.} We use $SD_{1.4}$ base model for our method, \textit{StableDiffusion}, and \textit{MultiDiffusion} and generate at the resolution $512 \times 680$. Similar to the horizontal resolutions, baseline methods exhibit several limitations such as poor text-image alignment ($1^{st}$ and $2^{nd}$ rows), repeated elements ($3^{rd}, 4^{th}$ and $5^{th}$ rows), and generated artifacts ($6^{th}$ and $7^{th}$ rows). In comparison, our method consistently maintains better image coherence and fidelity to the input prompts.}
    \label{fig:supp-portrait-comparison}
\end{figure*}

\begin{figure*}
    \centering
    \includegraphics[width=0.9\textwidth, height=0.9\textheight]{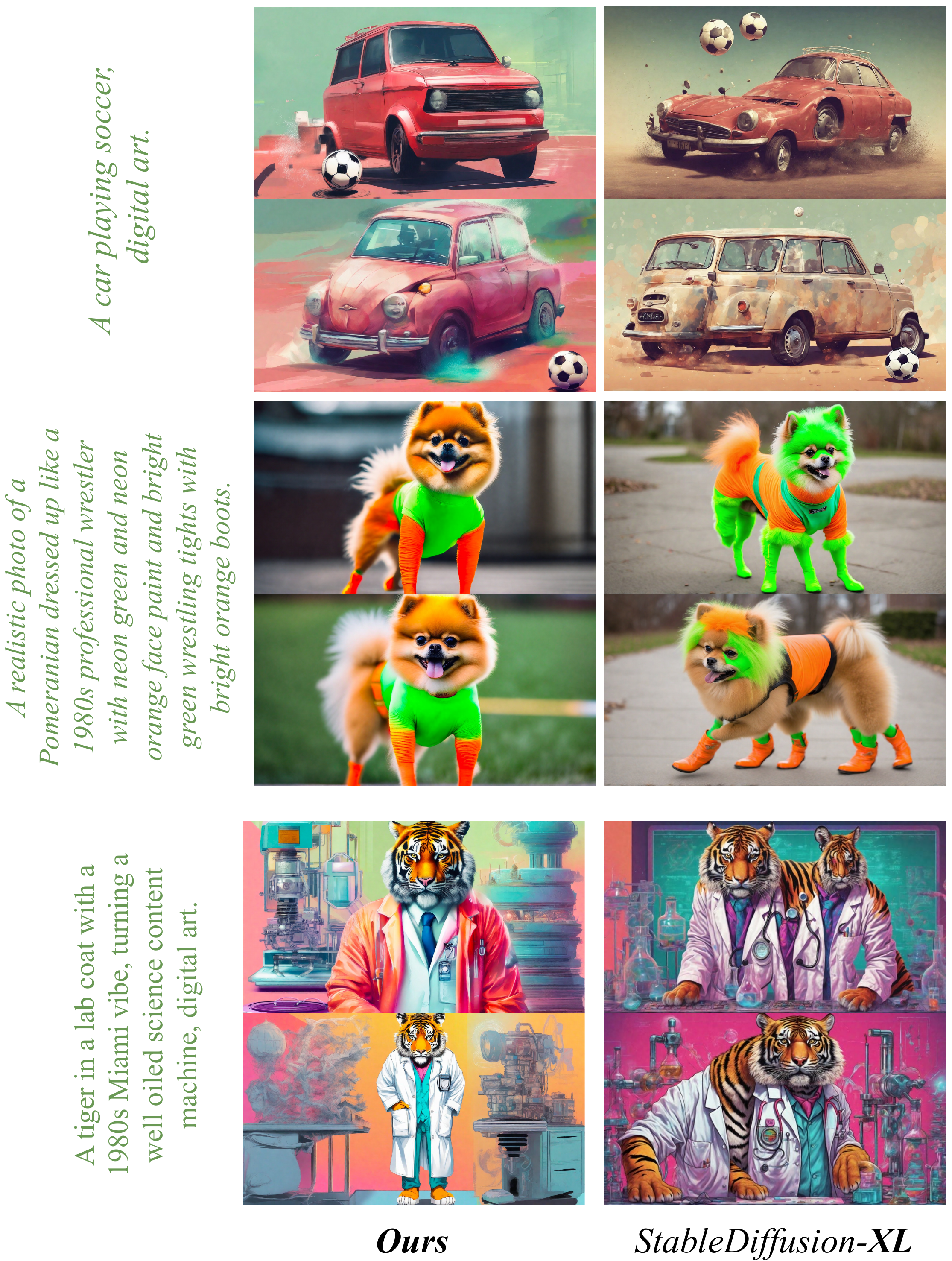}
    \caption{\textbf{Additional qualitative comparison with SDXL on Full-HD horizontal images}. We use randomly sampled DrawBench prompts from the Reddit Category. Despite its fine-tuning process, \emph{StabelDiffusion-XL} produces images with repetitive textures and elements in full-HD resolution. Our method achieves a more cohesive composition while maintaining a comparable level of details, all while requiring less memory.}
    \label{fig:supp-landscape-sdxl-comparison-1}
\end{figure*}

\begin{figure*}
    \centering
    \includegraphics[width=0.85\textwidth]{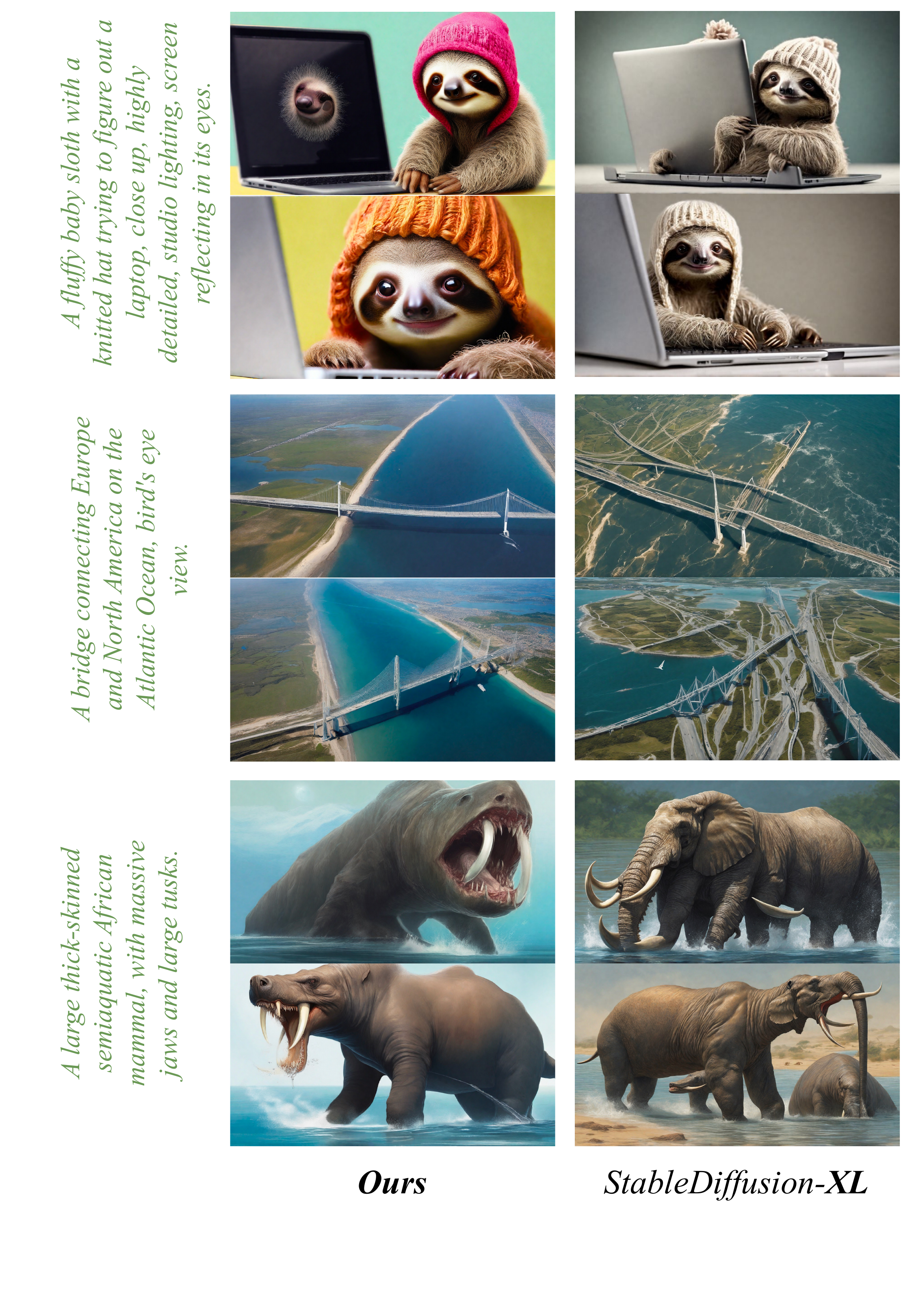}
    \caption{\textbf{Additional qualitative comparison with SDXL on Full-HD horizontal images}. We use randomly sampled DrawBench prompts from the Reddit Category. \emph{StabelDiffusion-XL} produce images that tend to repeat body parts and texture in full-HD resolution. In comparison, our method achieves better image coherence and maintains a similar perceptual quality, all while requiring less memory.}
    \label{fig:supp-landscape-sdxl-comparison-2}
\end{figure*}

\begin{figure*}
    \centering
    \includegraphics[width=\textwidth, height=0.9\textheight]{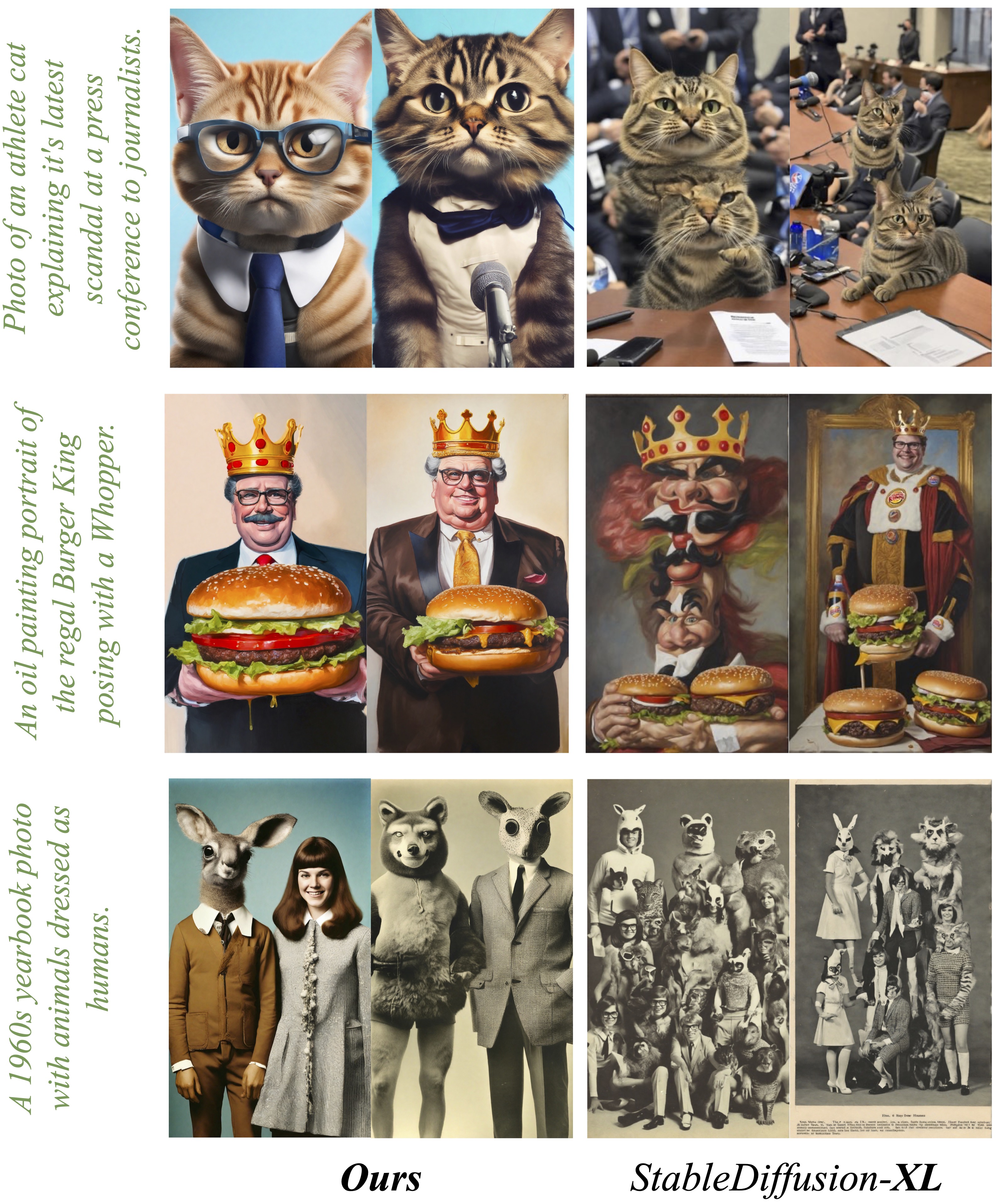}
    \caption{\textbf{Additional qualitative comparison with SDXL on Full-HD vertical images}. We use randomly sampled DrawBench prompts from the Reddit Category. Similar to horizontal resolutions,  \emph{StabelDiffusion-XL} produce repeated elements in full-HD resolution. In comparison, our method achieves more coherent images and generates content that fits the frame aspect ratio.}
    \label{fig:supp-portrait-sdxl-comparison-1}
\end{figure*}

\begin{figure*}
    \centering
    \includegraphics[width=\textwidth]{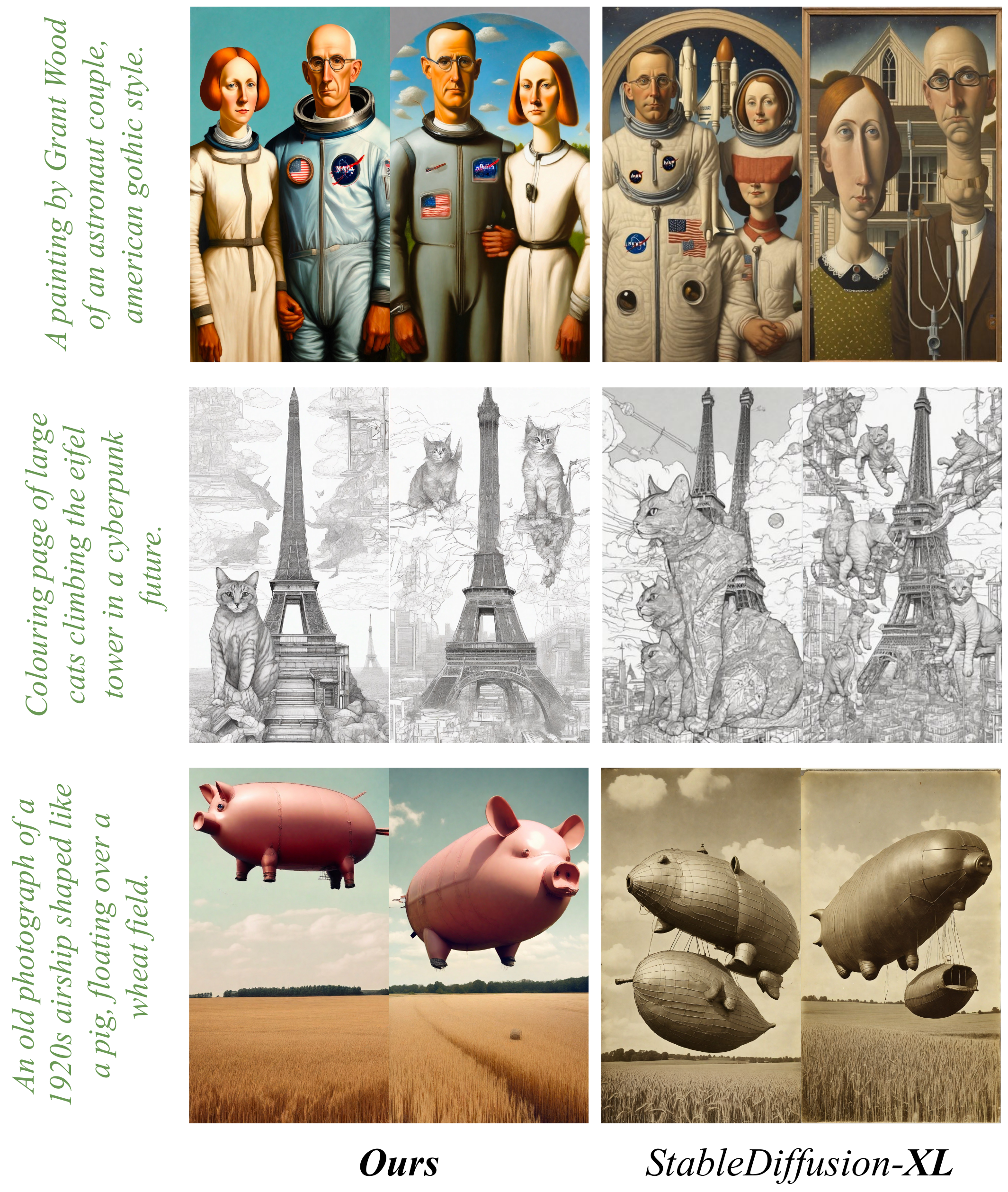}
     \caption{\textbf{Additional qualitative comparison with SDXL on Full-HD vertical images}. We use randomly sampled DrawBench prompts from the Reddit Category. Similar to horizontal resolutions,  \emph{StabelDiffusion-XL} produce repeated elements that significantly affect the image coherence. In comparison, our method achieves superior composition while maintaining a similar level of detail.}
    \label{fig:supp-portrait-sdxl-comparison-2}
\end{figure*}

\begin{figure*}
    \centering
    \includegraphics[width=\textwidth]{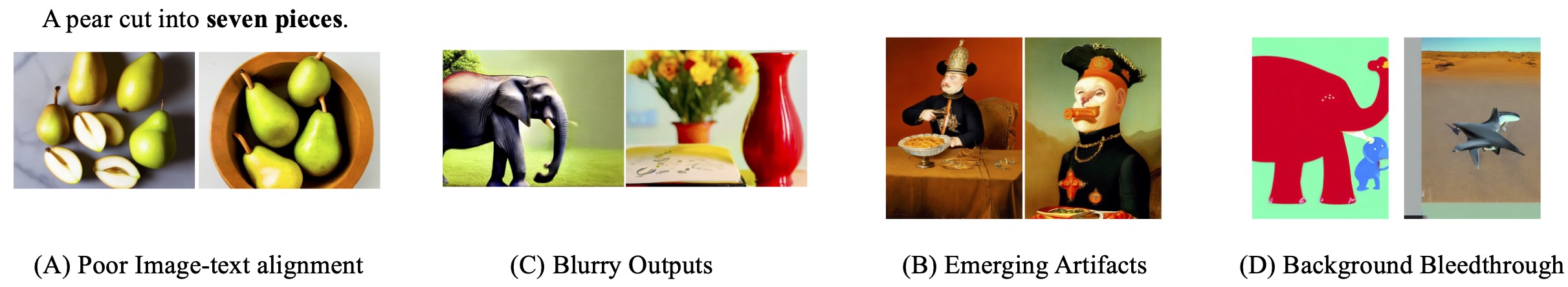}
    \caption{\textbf{Limitations of \name.} (A) poor text-image alignment for complex prompts, inherited from the base diffusion model, (B) increased blur in outputs with higher RRG weight, (C) emerging artifacts in complex images, and (D) rare background bleed-through where the color-constant background is unintentionally included in the generated image.}
    \label{fig:supp-limitations}
\end{figure*}
}

\end{document}